\documentclass[10pt,twocolumn,letterpaper]{article}

\usepackage{cvpr}              

\usepackage[dvipsnames]{xcolor}

\usepackage{booktabs}
\usepackage{bm}
\usepackage{times}
\usepackage{epsfig}
\usepackage{float} 
\usepackage{makecell}
\usepackage{colortbl}
\usepackage{mathtools}
\usepackage{extarrows}
\usepackage{algorithmic}
\usepackage{adjustbox}
\usepackage[percent]{overpic}
\usepackage[linesnumbered,ruled]{algorithm2e}
\SetKwComment{Comment}{$\triangleright$\ }{}
\usepackage{balance}
\usepackage{stfloats}
\usepackage{textcomp}
\usepackage{subcaption}
\usepackage{pifont}

\usepackage{multirow}
\usepackage{comment}

\graphicspath{{iccv2023AuthorKit/figures/}}

\definecolor{cvprblue}{rgb}{0.21,0.49,0.74}
\definecolor{cYellow}{HTML}{FFFFCC}
\definecolor{cRed}{HTML}{FFCCCC} 
\definecolor{cGrey}{HTML}{F3F7F2} 
\definecolor{cGreen}{HTML}{339933}
\usepackage[pagebackref,breaklinks,colorlinks,citecolor=cvprblue]{hyperref}

\usepackage[dvipsnames]{xcolor}
\hypersetup{
    colorlinks=true,       
    linkcolor=red,          
    citecolor=cGreen,        
} 
\usepackage{colortbl}

\graphicspath{{figures/}}

\newcommand{\correspondingauthormark}{\textsuperscript{$\dagger$}}

\def\paperID{0579} 
\def\confName{CVPR}
\def\confYear{2024}

\title{Text Is MASS: Modeling as Stochastic Embedding for Text-Video Retrieval}

\author{Jiamian Wang$^{1}$,
Guohao Sun$^{1}$,
Pichao Wang$^{2}$\thanks{The work does not relate to author's position at Amazon.},
Dongfang Liu$^{1}$\correspondingauthormark,\\
Sohail Dianat$^{1}$,
Majid Rabbani$^{1}$,
Raghuveer Rao$^{3}$,
Zhiqiang Tao$^{1}$\thanks{Corresponding authors: Dongfang Liu (dongfang.liu@rit.edu) and Zhiqiang Tao (zhqiang.tao@rit.edu)}\correspondingauthormark\\
$^{1}$Rochester Institute of Technology, 
$^{2}$Amazon Prime Video, 
$^{3}$Army Research Laboratory
}

\begin{document}
\maketitle

\begin{abstract}
The increasing prevalence of video clips has sparked growing interest in text-video retrieval. 
Recent advances focus on establishing a joint embedding space for text and video, relying on consistent embedding representations to compute similarity. 
However, the text content in existing datasets is generally short and concise, making it hard to fully describe the redundant semantics of a video. Correspondingly, a single text embedding may be less expressive to capture the video embedding and empower the retrieval.
In this study, we propose a new stochastic text modeling method T-MASS, i.e., \underline{t}ext is \underline{m}odeled \underline{a}s a \underline{s}tocha\underline{s}tic embedding, to enrich text embedding with a flexible and resilient semantic range, yielding a text mass. To be specific, we introduce a similarity-aware radius module to adapt the scale of the text mass upon the given text-video pairs. Plus, we design and develop a support text regularization to further control the text mass during the training. 
The inference pipeline is also tailored to fully exploit the text mass for accurate retrieval. 
Empirical evidence suggests that T-MASS not only effectively attracts relevant text-video pairs while distancing irrelevant ones, but also enables the determination of precise text embeddings for relevant pairs.
Our experimental results show a substantial improvement of  T-MASS over baseline ($3\%\sim6.3\%$ by R@1). Also, T-MASS achieves state-of-the-art performance on five benchmark datasets, including MSRVTT, LSMDC, DiDeMo, VATEX, and Charades.  
{Code and models are available \href{https://github.com/Jiamian-Wang/T-MASS-text-video-retrieval}{here}}.    
\end{abstract}
    
\section{Introduction}
\label{sec:intro}

Text-video retrieval is to find the most semantically relevant video clip (text) from a candidate pool referring to the text (video clip) query~\cite{bain2021frozen,luo2022clip4clip,xue2022clip,bogolin2022cross,cheng2021improving,portillo2021straightforward}. 
Performing an accurate retrieval is non-trivial due to the divergent characteristics of video and text: videos tend to offer redundant semantic clues~\cite{zhao2022centerclip,fang2023uatvr,gorti2022x}, inevitably posing challenges for the feature extraction, while the text, commonly appearing as short captions, subtitles, and even hashtags, seems to be semantically limited by comparison to videos~\cite{lin2022text}.

\begin{figure}[t] 
\centering 
\includegraphics[width=.475\textwidth]{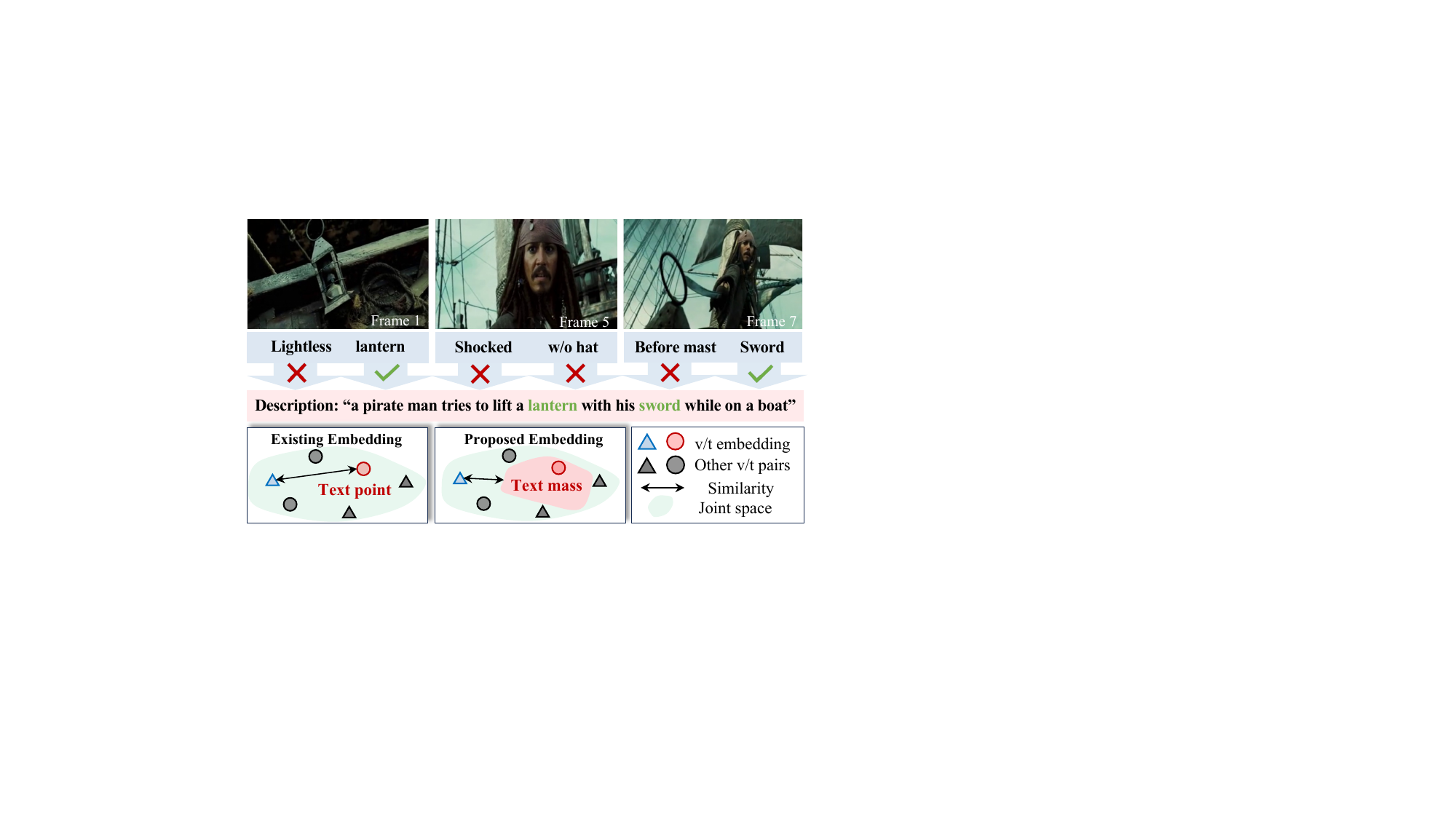}
\vspace{-6mm}
\caption{ 
Text inside a relevant video is hard to fully describe the redundant semantics of the video. Correspondingly, single text embedding may be less expressive to handle the video information in joint space. We propose a new embedding of text mass with a resilient semantic range, to better capture rich video clues. 
} 
\label{fig: coverfig}
\vspace{-5mm} 
\end{figure}

Recognizing the nature of video and text, some prior works~\cite{luo2022clip4clip,xue2022clip} adapt  powerful vision-language models (e.g., CLIP~\cite{radford2021learning}) to the multimodal domain~\cite{bain2022clip} of text and video. Others learn enhanced video representation through video-text interaction~\cite{gorti2022x,lin2022text,wang2022disentangled} or temporal modeling~\cite{bain2021frozen,liu2022ts2}. Besides, bridging video and text at a fine granularity
also forms a promising direction~\cite{wang2021dig,guan2023pidro,wang2021t2vlad,zhu2020actbert,jin2023video,jin2023text}. 
In summary, prevailing text-video retrieval methods are mainly dedicated to extracting accurate video or text embedding, such as text/video points, for retrieval.

Despite the success of video/text embedding methods, it is  hard to learn a single text embedding to fully cover all the semantics and visual variations inside a video, since the text content  is usually short and concise, which contains limited semantics compared with its paired relevant video (see Fig.~\ref{fig: coverfig} \textit{top}). 
This fact exacerbates alignment difficulties,
where text may not adequately express the richness of video information.
Drawing inspiration, we  provide a more flexible text modeling approach to capture rich video semantic clues, thus enabling a better alignment between video and text semantics. We introduce T-MASS, \emph{i.e.},  \textbf{\underline{T}}ext is   \textbf{\underline{M}}odeled \textbf{\underline{A}}s a \textbf{\underline{S}}toch\textbf{\underline{S}}tic embedding. Unlike existing methods, the proposed method no longer treats text as a single point in the embedding space, but projects it as a ``mass'' (see Fig.~\ref{fig: coverfig} \textit{bottom}) to enable a resilient semantic range to account for the potential misalignment between video and text embeddings.  A straightforward way to implement text mass can adopt the reprametrization~\cite{kingma2013auto} upon the deterministic text embedding given by CLIP. However, learning such a text mass imposes the following challenges.

First, it is non-trivial to determine the scale of the text mass. The underlying scale is text-dependent and can even be dynamic relative to different videos. To this end, we develop a similarity-aware radius module to enable a learnable scale adaptive to text-video pairs. 
Second, how to further regularize and shift the text mass in the joint embedding space is an open question. Without jointly processing the whole text mass, we find that solely performing contrastive learning between sampled stochastic text points and video points during training brings promising performance. Besides, we locate a support text vector upon the text mass, taking it as a proxy to simultaneously control the position and scale of the text mass relative to the query video. 
We also reformulate the inference process to fully exploit the text mass for more effective text-video retrieval. For each video candidate, we first sample a batch of stochastic text embedding for the query text and choose the closest one to the video embedding for the evaluation. 
Interestingly, we find that the proposed T-MASS not only bridges the relevant pairs and pushes the irrelevant ones (Fig.~\ref{fig: postive-negative}), compared with the single-point text representation, but also empowers a precise text semantics mapping (Fig.~\ref{fig: dynamic-R-1}). 
We summarize the contributions of this work as follows.

\begin{itemize}
    \setlength\itemsep{0em}
    \item 
    This work rethinks the design of text embedding for text-video retrieval. 
    We propose T-MASS as a new stochastic modeling approach to enable expressive and flexible text embedding to better capture video clues and align text and video semantics in joint space.
    
    \item 
    This work provides a representative design of similarity-aware radius network to encourage the text semantics  alignment, facilitating a resilient and flexible text embedding that can adapt to the video variations. 

    \item 
    This work develops an effective learning strategy upon stochastic text embedding, 
    specifically, a stochastic symmetric cross-entropy learning objective to learn an effective text mass. Besides, a support text  vector as a regularization  to further scale and shift the text mass. 

    \item 
    The proposed method improves the baseline by a large margin ($+3\%\sim6.3\%$ at R@1),  setting the new state-of-the-art on five benchmark datasets, including MSRVTT, LSMDC, DiDeMo, Charades, and VATEX. Extensive analyses find that T-MASS not only better distances irrelevant pairs and attracts relevant pairs, but also enables a more promising text semantics learning for relevant pairs. 

\end{itemize}
\section{Related Work}
\label{sec:Related Work}

\textbf{Text-video Retrieval}.
JSFusion~\cite{yu2018joint} pioneered the exploration of hierarchical similarities between video and text using a convolutional decoder, establishing a benchmark for the task. Transformer~\cite{vaswani2017attention,dosovitskiy2020image}-based methods~\cite{gabeur2020multi,dzabraev2021mdmmt,CLIP2TV,Li_2023_CVPR,10205280,10204414,deng2023prompt} abstract multi-modal data clues via cross attention, resulting in significant performance gains. Recent advancements leverage CLIP~\cite{radford2021learning}  for the semantics extraction~\cite{gorti2022x,luo2022clip4clip,xue2022clip,wu2023cap4video,xu-etal-2021-videoclip,zhao2022centerclip,lei2021less}, \emph{e.g.}, CLIP4Clip~\cite{luo2022clip4clip}  discusses the transferability of pre-trained CLIP model to text-video retrieval. To solve the domain gap,  CLIP-ViP~\cite{xue2022clip} exploits the video post-pretraining, achieving state-of-the-art results. Alternatively, Cap4Video~\cite{wu2023cap4video} harnesses the power of a pre-trained large model by introducing additional captions, bringing insight on fully taking advantage of augmented data. TEACHTEXT~\cite{croitoru2021teachtext} empowers the retrieval by leveraging multiple text encoders.  Besides, DiffusionRet~\cite{jin2023diffusionret} advances by integrating diffusion model into the text-video retrieval. 
Additionally, introducing additional modality, \emph{e.g.}, audio~\cite{ibrahimi2023audio, NEURIPS2021_cb3213ad,lin2022eclipse,miech2018learning,liu2022animating}, draws increasing attention. 
The proposed method opts to learn a expressive and powerful text embedding, achieving substantial improvements without post-pretrain the CLIP with additional video data. The proposed method can even outperform previous methods enhanced by post-processing techniques~\cite{bogolin2022cross,cheng2021improving}.

\textbf{Text and Video Representation Learning}. This work builds upon CLIP~\cite{radford2021learning} model, attributing to its promising semantics extraction. Based on CLIP, existing methods predominantly focus on enhanced video and text representations for retrieval~\cite{fang2021clip2video,gorti2022x,guan2023pidro,li2023progressive, 10203496,ADVMM,Liu2019UseWY,10209009,PRVR,jin2022expectation}. 
TS2-Net~\cite{liu2022ts2} models fine-grained temporal visual clues, showcasing promising performance. X-Pool~\cite{gorti2022x} exploits text-conditioned feature fusion across frames, delivering more semantically similar embedding. PIDRo~\cite{guan2023pidro} and ProST~\cite{li2023progressive} model the informative semantic clues of video and text in a fine-grained manner, achieving encouraging performances. UATVR~\cite{fang2023uatvr} innovatively recognizes  and models uncertainties in both modalities. In contrast to these methods representing text and video embedding with a common form, 
T-MASS implements a stochastic text embedding, jointly learning a text mass and a video point in embedding space. Notably, the proposed method achieves remarkable performance boost without requiring sophisticated designs on video feature extraction, such as temporal modeling~\cite{bain2021frozen,liu2022ts2, 10205049}, fine-grained alignment~\cite{wang2021t2vlad,chen2020fine,LUNVR}, etc.

\section{Method}

\begin{figure*}[h] 
\centering 
\includegraphics[width=\textwidth]{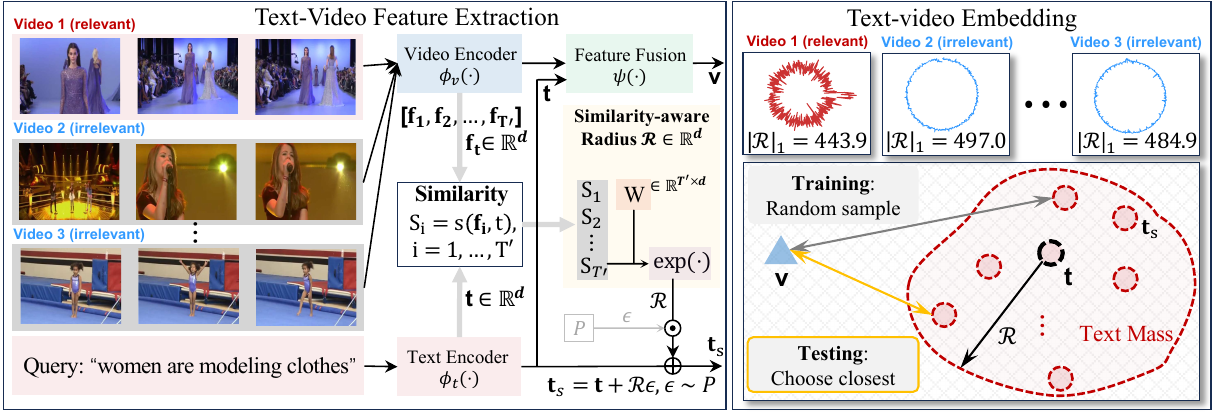}
\vspace{-0.6cm}
\caption{
Illustration of the proposed text-video retrieval method T-MASS, which adopts dual-branch CLIP~\cite{radford2021learning} ($\phi_v$ and $\phi_t$) to extract frame features $[\mathbf{f}_1,...,\mathbf{f}_{T'}]$ and text embedding $\mathbf{t}$. Then a feature fusion module $\psi$ is employed to produce video embedding $\mathbf{v}$. We 
develop a similarity-aware module $\mathcal{R}$ to facilitate the  reparameterization~\cite{kingma2013auto} of the stochastic text embedding $\mathbf{t}_s$, yielding a text mass in the joint space. During training, we compute the loss upon $\mathbf{v}$ and random sampled $\mathbf{t}_s$. During evaluation, we collect a group of $\mathbf{t}_s$ and select the one exhibiting the highest similarity with $\mathbf{v}$. We visualize the learned radius $\mathcal{R}$ for relevant/irrelevant pairs. 
More details are in Section~\ref{subsec: T-MASS}.
 }
\label{fig: framework}
\vspace{-0.2cm}
\end{figure*}

\subsection{Preliminaries}\label{subsec: vtr}

We denote the text as $t$ and the raw video clip as $v$. The task of text-video retrieval firstly involves learning embedding for text and video in a joint space, yielding $\textbf{t}, \textbf{v} \in \mathbb{R}^d$,  where $d$ represents the feature dimension. A similarity measuring function $s(\textbf{t}, \textbf{v})$, such as cosine similarity, is then employed to compute relevancy. Given a training dataset with $K$ different text-video pairs, $\mathcal{D}=\{(t_k, v_k)\}^{k=K}_{k=1}$, a widely-used loss function for this task can be a symmetric cross entropy~\cite{oord2018representation}, which minimizes the distance of relevant text-video pairs while maximizes distance of irrelevant pairs. Both text-to-video ($t \rightarrow v$) and video-to-text ($v \rightarrow t$) are considered in this approach by
\begin{equation}\label{eq: symmetric ce}
\begin{aligned}
    \mathcal{L}_{t \rightarrow v} = - \frac{1}{N}\sum\limits^{N}_{i=1}\log\frac{e^{s(\mathbf{t}_i, \mathbf{v}_i)\cdot \lambda}}{\sum\nolimits_{j}e^{s(\mathbf{t}_i, \mathbf{v}_j)\cdot \lambda}}, \\
    \mathcal{L}_{v \rightarrow t} = - \frac{1}{N}\sum\limits^{N}_{i=1}\log\frac{e^{s(\mathbf{t}_i, \mathbf{v}_i)\cdot \lambda}}{\sum\nolimits_{j}e^{s(\mathbf{t}_j, \mathbf{v}_i)\cdot \lambda}}, 
\end{aligned}
\end{equation}
where $N$ is a collection of text-video pairs, typically representing the batch size, and $\lambda$ is a learnable scaling factor. Text and video embedding, $\mathbf{t}_i$ and $\mathbf{v}_i$/$\mathbf{v}_j$, are produced by delicate feature extractors with learnable parameters.  The overall loss function $\mathcal{L}_\text{ce}$ is 
\begin{equation}\label{eq: contrastive loss}
\begin{aligned}
    \mathcal{L}_\text{ce}=\frac{1}{2} (\mathcal{L}_{t \rightarrow v} + \mathcal{L}_{v \rightarrow t}). 
\end{aligned}
\end{equation}
The loss function reaches zero when all text-video pairs in a batch is entirely relevant, \emph{i.e.}, $s(\mathbf{t}_i, \mathbf{v}_i)=1$, and $s(\mathbf{t}_i, \mathbf{v}_j)=0$, $i\neq j$, for all irrelevant pairs. This is non-trivial and highly depends on the quality of text and video embedding ($\mathbf{t}$ and $\mathbf{v}$).  
As shown in Fig.~\ref{fig: coverfig}, in practice, even text-video pairs that are identified as ``relevant'' could be not entirely consistent -- video $v$ provides redundant clues, while text $t$ may contain limited semantics. This poses challenges to the semantics extraction for both modalities.

\subsection{Text-Video Representations}\label{subsec: previous methods}

\noindent\textbf{Feature Extraction}.
The extraction of multi-modal semantic embedding has gained much attention~\cite{gabeur2020multi,yu2018joint}. Recent advancement of CLIP~\cite{radford2021learning} in recent text-video retrieval methods~\cite{xue2022clip,gorti2022x}, and thus we primarily focus on CLIP-based methods in this work. Given a video comprising $T$ frames, denoted as $v=[f_1, ..., f_T]$, the widely-used protocol is to sample $T'$ frames and feed them into CLIP, producing $T'$ different frame embedding $\mathbf{f}_i$, $i={1,...,T'}$. Let $\phi_v$ and $\phi_t$ denote the CLIP's image and text encoders. The feature extraction is given by 
\begin{equation}\label{eq: CLIP}
\begin{aligned}
     \mathbf{f}_i = \phi_v(f_i), i={1,...,T'};  ~~ \mathbf{t} = \phi_t(t), 
\end{aligned}
\end{equation}
where $\mathbf{f}_i\in\mathbb{R}^d$. Based on the frame embedding $[\mathbf{f}_1, ..., \mathbf{f}_{T'}]$, previous works develop various strategies to compute the final video embedding $\mathbf{v}$ for the similarity measurement
\begin{equation}\label{eq: f2v}
\begin{aligned}
    \mathbf{v} = \psi([\mathbf{f}_1, ..., \mathbf{f}_{T'}], \mathbf{t}), 
\end{aligned}
\end{equation}
where $\psi(\cdot)$ is the feature fusion module that abstracts the video semantics through frame-text interaction under different granularities, or relies on temporal modeling, etc. Despite the promising performance, 
it seems that finding a close alignment between $\mathbf{t}$ and $\mathbf{v}$ is still challenging since this requires an effective $\phi_v(\cdot)$ and $\psi(\cdot)$ for the video embedding learning, also calls for a powerful $\phi_t(\cdot)$ for text embedding determination. 
This motivates us to re-examine the text and video embedding as follows. 

\textbf{Motivation}. 
Existing methods emphasize more on learning video embedding $\mathbf{v}$, including frame sampling protocol, feature extraction $\phi_v(\cdot)$, and fusion $\psi(\cdot)$ designs, but pay less attention to text embedding $\mathbf{t}$. As shown in Section~\ref{sec:intro} and Fig.~\ref{fig: coverfig}, text $t$ is hard to fully describe the semantics of a video $v$, which yields $\mathbf{t}$ with less expressiveness and semantic clues to align with $\mathbf{v}$ in joint space. We are motivated to enhance the text embedding with more resilience and flexibility. 
Specifically, rather than using a single point, text embedding can be associated with a specific semantics \textit{range} that is resilient enough to incorporate (or be close to) the relevant video embedding. This leads us to introduce a new embedding called text mass,  setting it apart from existing methodologies.

\subsection{Proposed Method: T-MASS}\label{subsec: T-MASS}
\textbf{Stochastic Text Modeling}.
\label{subsubsec: network design}
In this work, we introduce T-MASS, \emph{i.e.}, \textbf{\underline{T}}ext is   \textbf{\underline{M}}odeled \textbf{\underline{A}}s a \textbf{\underline{S}}tocha\textbf{\underline{S}}tic representation. In contrast to prevalent treatments, T-MASS projects text as a ``mass'' to encourage expressive and resilient representations, jointly learning upon text-video embedding with distinct forms.  
Fig.~\ref{fig: framework} provides the framework of the proposed T-MASS. 
Specifically, we adopt reparameterization~\cite{kingma2013auto} to enable stochastic gradient calculations during the training. Based on the text embedding $\mathbf{t}$ given in Eq.~\eqref{eq: CLIP}, we introduce stochastic text embedding $\mathbf{t}_s\in\mathbb{R}^d$ as
\begin{equation}\label{eq: stochastic text}
\begin{aligned}
    \mathbf{t}_s =  \mathbf{t} + \mathcal{R}\cdot\mathbf{\epsilon}, \mathbf{\epsilon} \sim P, 
\end{aligned}
\end{equation}
where $\mathbf{\epsilon}$ is an auxiliary variable sampled from a prior distribution, \emph{e.g.}, $P=\mathcal{N}(\mathbf{0},\mathbf{1})$. $\mathcal{R}\in\mathbb{R}^d$ models the scale of the text mass and defines its underlying ``radius''.  Unlike previous embedding that aims to appropriately adjust the distance between two points, \emph{i.e.}, $\mathbf{t}$ and $\mathbf{v}$, any point that falls into the text mass is considered as a valid representation corresponding to the content of text $t$ and can be used for the similarity calculation, \emph{e.g.}, $s(\mathbf{t}_s, \mathbf{v})$.  By this means,  existing powerful text encoders can be naturally adopted, and minimum adjustments in implementation is required. 

\textbf{Similarity-Aware Radius Modeling}. 
Directly incorporating stochastic text embedding in Eq.~\eqref{eq: stochastic text} into the loss $\mathcal{L}_\text{ce}$ in Eq.~\eqref{eq: contrastive loss} is challenging.  On the one hand, an oversized text mass ($\mathcal{R}$ is too large) might improperly encompass (or improperly be close to) less relevant or even irrelevant video embedding points in the joint space, thus misleading the retrieval, On the other hand, too small text mass ($\mathcal{R}\rightarrow0$) may lacks expressiveness to bridge the video. 
Thereby, it is non-trivial to manually determine an optimal value for $\mathcal{R}$; rather, the underlying radius $\mathcal{R}$ should adapt to different text-video pairs. We propose a similarity-aware radius module to learn proper text mass scaling by jointly taking text $\mathbf{t}$ and video frames $[\mathbf{f}_1,...,\mathbf{f}_{T'}]$ before feature fusion as inputs. 

\begin{figure}[t]
\scriptsize
\hspace{-1.5mm}
\resizebox{1\columnwidth}{!}{
\begin{tabular}{cc}
\begin{adjustbox}{valign=t}
    \begin{tabular}{c}
    \includegraphics[width=0.85\columnwidth,angle=0]{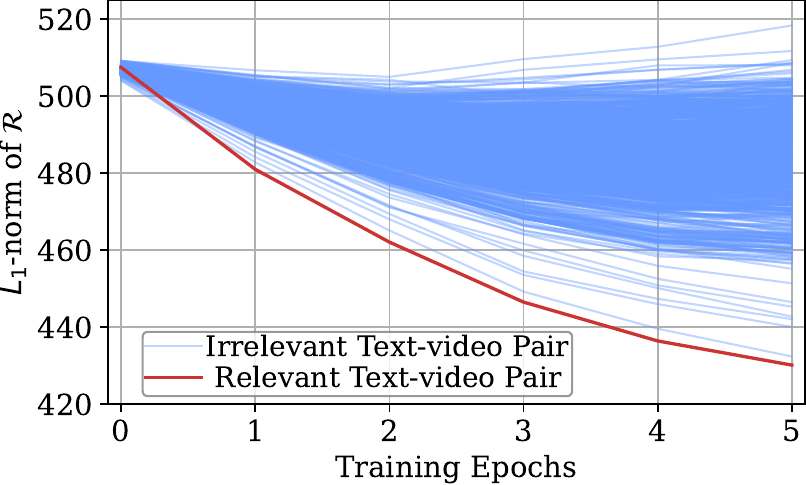}
    \end{tabular}
\end{adjustbox}
\hspace{-2mm} 
\begin{adjustbox}{valign=t}
    \begin{tabular}{c}
    \includegraphics[width=0.2\columnwidth,angle=0]{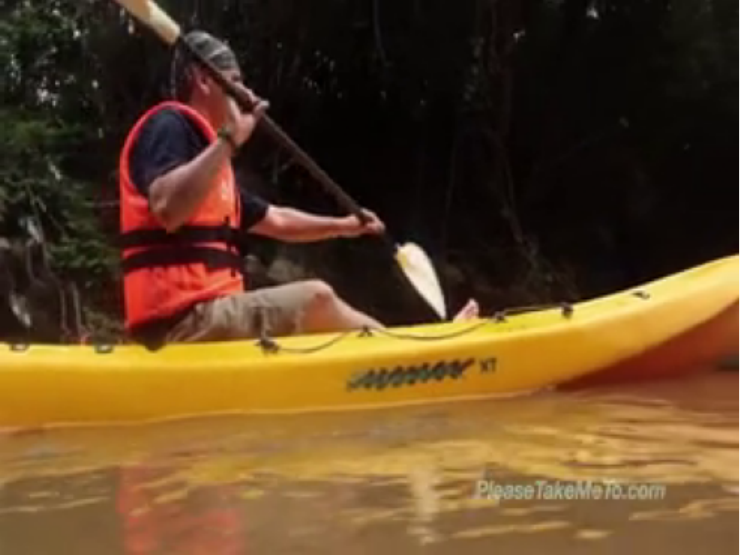} \\ 
    \includegraphics[width=0.2\columnwidth,angle=0]{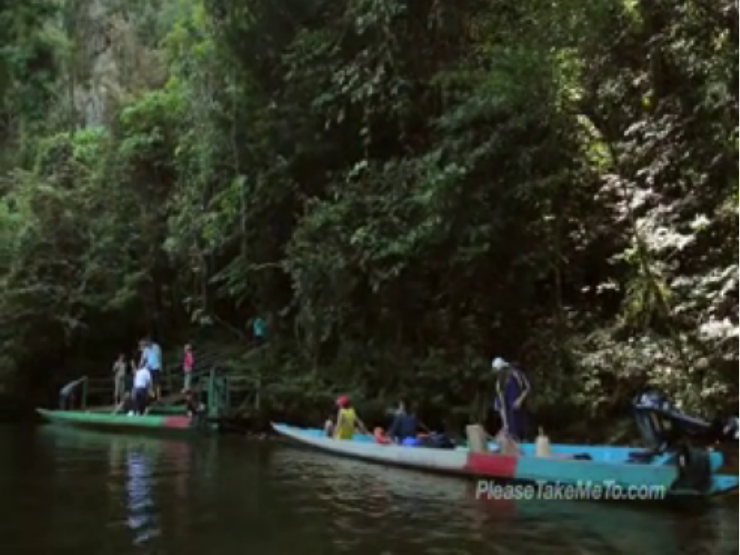} \\
    \includegraphics[width=0.2\columnwidth,angle=0]{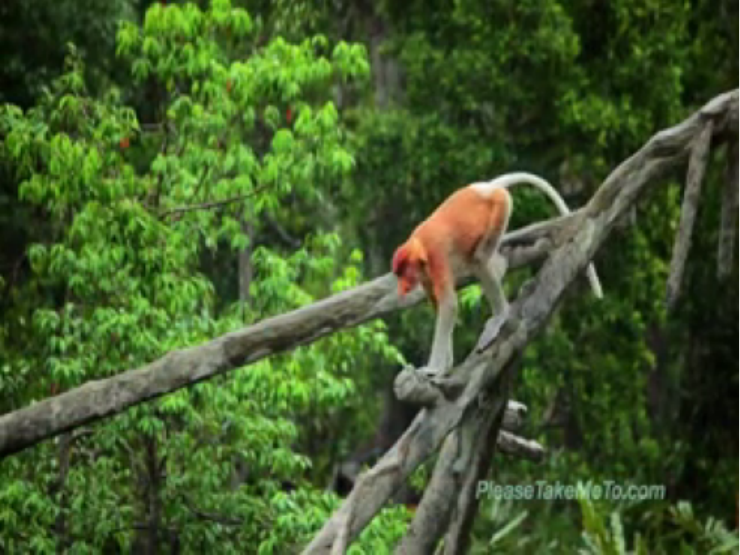} \\
    \end{tabular}
\end{adjustbox}
\end{tabular}
}
\vspace{-2.5mm}
\caption{
Dynamics of $\mathcal{R}$. 
We plot $|\mathcal{R}|_1$ for a relevant $t$-$v$ pair  (130-th in MSRVTT-1K, video on the \textit{right}) and the query text with $999$ irrelevant videos. T-MASS learns a precise text semantics for the relevant pair (smallest $|\mathcal{R}|_1$). This is  typically observed on correctly retrieved pairs. More examples are in supplementary.
}
\label{fig: dynamic-R-1}
\vspace{-5mm}
\end{figure}

The key idea is to first compute the cosine similarity of a text-video pair and leverage it as an indicator of the text-video relationship. For 
instance, if the text-video pair exhibits relevance, 
we expect a well-aligned text mass with a proper radius and position that potentially allows an accurate retrieval, as shown by the red curve in Fig.~\ref{fig: dynamic-R-1}. Reversely, it is less likely to learn a meaningful mass when they are irrelevant, \emph{e.g.}, imprecise text mass (blue curves in Fig.~\ref{fig: dynamic-R-1}). Given the embedding $\mathbf{t}$ and frame embedding $[\mathbf{f}_1, ..., \mathbf{f}_{T'}]$ by Eq.~\eqref{eq: CLIP}, we compute the text-video similarity as  
\begin{equation}\label{eq: linearcos-sim}
\begin{aligned}
    S_i = s(\mathbf{t}, \mathbf{f}_i), i=1,...,T', 
\end{aligned}
\end{equation}
based on which we propose a learnable scalar $\theta$ to compute the radius
    $\mathcal{R} = \text{exp}(\frac{\theta}{T'}\sum\nolimits^{T'}_{i=1}S_i)$
where $\theta$ is broadcast to fit the dimension $d$. We use an exponential function to scale the radius further. We observe that such an implementation has brought an encouraging performance boost, compared with the unlearnable strategy, \emph{i.e.}, defining $\mathcal{R}$ purely upon similarities values given by Eq.~\eqref{eq: linearcos-sim}. Besides, solely using a scalar to adjust the radius in a high-dimensional space might be less flexible. As shown in Fig.~\ref{fig: framework}, we also explore a linear layer to compute $\mathcal{R}$ by
\begin{equation}\label{eq: linearproj}
\begin{aligned}
    \mathcal{R} = \text{exp}(\mathbf{S}\mathbf{W}), \mathbf{S}=[S_1, ..., S_{T'}],  
\end{aligned}
\end{equation}
where $\mathbf{W}\in\mathbb{R}^{T'\times d}$ denotes the learnable weights in the linear layer. 
The resulting $\mathcal{R}$ will take effect in the stochastic text embedding calculation as shown in Eq.~\eqref{eq: stochastic text}. In Section~\ref{subsec: discussion}, we provide the corresponding discussion  toward the design principles of $\mathcal{R}$. See more details in Table~\ref{tab: R ablation}.

\textbf{Learning Text Mass in Joint Space}. \label{subsubsec: learning obj}
The original loss function in Eq.~\eqref{eq: symmetric ce} between $\mathbf{t}$ and $\mathbf{v}$ may only take effect on shifting the text mass without controlling its scale. Since the text mass is implemented by stochastic text embedding, we randomly sample a stochastic text embedding $\mathbf{t}_s$ and use it to replace the text embedding $\mathbf{t}$ in Eq.~\eqref{eq: symmetric ce} during training, so that different points in text mass participate into the learning.  Distinguished from the original symmetric cross-entropy loss $\mathcal{L}_\text{ce}$, we denote this stochastic loss  $\mathcal{L}_{s}$. The overall loss function becomes $\mathcal{L}_\text{total}=\mathcal{L}_\text{ce} + \mathcal{L}_s$. We show that such a learning schedule brings notable benefits (\emph{e.g.}, $>1.5\%+$ at R@1). Moreover, we find that specifically regularizing over $\mathbf{t}$ (\emph{i.e.}, using $\mathcal{L}_\text{ce}$) during the learning is unnecessary and can even be harmful. As, compared to $\mathbf{t}_s$,  $\mathbf{t}$ cannot reflect the context and position of the text mass, thus focusing on $\mathbf{t}$ can lead to a biased text mass learning. 
In addition, given that the text mass presents as an irregular volume in a complex and high-dimensional embedding space, learning a limited amount of $\mathbf{t}_s$ points seems insufficient to regularize the whole text mass. A straightforward solution is to introduce KL-divergence to further control the scale. However, it is challenging to determine an optimal prior.

\begin{figure}[t] 
\centering 
\includegraphics[width=.46\textwidth]{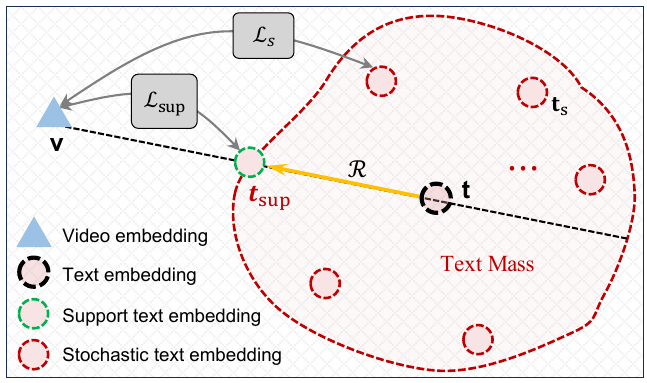}
\vspace{-3mm}
\caption{
Support text regularization. Besides computing the loss between the video embedding $\mathbf{v}$ and stochastic text embedding $\mathbf{t}_s$, we identify a support text embedding locating along the direction from $\mathbf{v}$ to $\mathbf{t}$ and being placed at the surface of the text mass, which serves as a proxy to enable text mass shifting and scaling. } 
\label{fig: support loss}
\vspace{-5mm} 
\end{figure}

We propose to identify a support text embedding vector located at the border of the text mass as a proxy to help adjust the scale and the shift of the text mass, termed as support text vector $\mathbf{t_\text{sup}}$. As shown in Fig.~\ref{fig: support loss},  $\mathbf{t_\text{sup}}$ serves as a stochastic text embedding sample locating along the direction from $\mathbf{v}$ to $\mathbf{t}$ and being placed at the surface of the text mass.
Therefore, pulling $\mathbf{v}$ and $\mathbf{t_\text{sup}}$ together or pushing them away to a large extent can help manipulate the text mass. We compute $\mathbf{t_\text{sup}}$ based on $\mathbf{t}$ given in Eq.~\eqref{eq: CLIP}, video embedding $\mathbf{v}$ upon Eq.~\eqref{eq: f2v}, and the radius modeling $\mathcal{R}$ as
\begin{equation}\label{eq: support text}
\begin{aligned}
    \mathbf{t_{\text{sup}}} = \mathbf{t} + \frac{\mathbf{v} - \mathbf{t}}{|\mathbf{v} - \mathbf{t}|}\mathcal{R}, 
\end{aligned}
\end{equation}
based on which we introduce another contrastive loss term with the same formulation as Eq.~\eqref{eq: contrastive loss} but only exchange $\mathbf{t}$ with $\mathbf{t_\text{sup}}$. We denote this regularization as $\mathcal{L}_\text{sup}$. During training, we not only sample a stochastic text embedding to compute the contrastive loss with the video, but also constantly pay attention to the support text vector. Our experiment in Section~\ref{subsec: discussion} shows that such a regularization brings a remarkable performance boost. 
The resulting loss function of the proposed method is given by
\begin{equation}\label{eq: overall loss}
\begin{aligned}
    \mathcal{L}_\text{total}  = \mathcal{L}_s + \alpha \mathcal{L}_\text{sup},
\end{aligned}
\end{equation}
where $\alpha$ is the support text regularization weight.
In Section~\ref{subsec: discussion} and Table~\ref{tab: ablation-loss}, we provide a complete comparison of different learning strategies. Besides, the proposed learning strategy encourages a better alignment for relevant/irrelevant text-video pairs\footnote{See supplementary material for more discussions about stochastic text embedding and KL-divergence .}. See Fig.~\ref{fig: postive-negative}  for more details.

\textbf{Inference pipeline}.
Building upon the proposed stochastic text representation, we modify the inference pipeline to take advantage of text mass. For any given text-video pairs $\{t, v\}$, we first extract text and frame features, $\mathbf{t}$ and $[\mathbf{f}_1, ..., \mathbf{f}_{T'}]$ using Eq.~\eqref{eq: CLIP}. Subsequently, we conduct $M$ times stochastic sampling upon Eq.~\eqref{eq: stochastic text}, producing $\{\mathbf{t}^1_s,...,\mathbf{t}^M_s\}$. We then select an optimal text embedding that gives the highest similarity with the video by
\begin{equation}\label{eq: text selection}
\begin{aligned}
    \widehat{\mathbf{t}}_s = \mathop{\arg\max}\limits_{\mathbf{t}_s} ~s(\mathbf{t}^i_s, \mathbf{v}), ~~i = 1,...,M, 
\end{aligned}
\end{equation}
where $\mathbf{v}$ is computed by the feature fusion module $\psi(\cdot)$ according to Eq.~\eqref{eq: f2v}. $\widehat{\mathbf{t}}_s$ is the final text embedding selected from the text mass for the metric computation. This strategy ensures that text embedding linked to the input text $t$ is no longer fixed and adaptive to videos, which holds benefits for retrieval by exploring more possibilities of the text embedding (especially, the ones that are closer to the video than the original $\mathbf{t}$). Note that this strategy is applicable for both text-to-video and video-to-text retrievals.

In summary, we introduce T-MASS, a stochastic text modeling method for  for text-video retrieval. Diverging from existing methods, T-MASS advances the retrieval by empowering text embedding with more expressiveness and flexibility. Besides the encouraging performance, T-MASS enables a better text-video alignment and text semantics adaptation. See  detailed illustrations and analysis below.

\section{Experiment}
\label{sec:experiment}

\subsection{Experimental Settings}
\label{subsec: setting}

\begin{table*}[t]
\begin{center}
\scalebox{0.96}{
\begin{tabular}{c|ccccc|ccccc} 
    \hline 
     \multirow{2}{*}{Method} & \multicolumn{5}{c|}{MSRVTT Retrieval} & \multicolumn{5}{c}{LSMDC Retrieval} \\
     \cline{2-11}
     & R@1 \small{$\uparrow$} & R@5 \small{$\uparrow$} & R@10 \small{$\uparrow$} & MdR \small{$\downarrow$} & MnR \small{$\downarrow$} & R@1 \small{$\uparrow$} & R@5 \small{$\uparrow$} & R@10 \small{$\uparrow$} & MdR  \small{$\downarrow$} & MnR \small{$\downarrow$} \\
    \hline
    {\color{cGreen}{\color{cGreen}\textit{\small{CLIP-ViT-B/32}}}} & & & & & & & & & & \\ 
    X-Pool~\cite{gorti2022x}  & 46.9 & 72.8 & 82.2 & 2.0 & 14.3 & 25.2 & 43.7 & 53.5 & 8.0 & 53.2  \\
    DiffusionRet~\cite{jin2023diffusionret}   & 49.0 & 75.2 & 82.7 & 2.0 & 12.1 & 24.4 & 43.1 & 54.3 & 8.0 & \textbf{40.7} \\
     UATVR~\cite{fang2023uatvr}  & 47.5 & 73.9 & 83.5 & 2.0 & 12.3  & -- & -- & -- & -- & -- \\
    TEFAL~\cite{ibrahimi2023audio} &49.4 & \textbf{75.9} & 83.9& 2.0& 12.0& 26.8 &46.1 &56.5 &7.0 &44.4 \\
    CLIP-ViP~\cite{xue2022clip} & 50.1 & 74.8 & 84.6 & 1.0 & -- & 25.6 & 45.3 & 54.4 & 8.0 & --\\
    \cellcolor{cGrey}T-MASS (\small{Ours})  &\cellcolor{cGrey}\textbf{50.2} &\cellcolor{cGrey}75.3
    &\cellcolor{cGrey}\textbf{85.1} &\cellcolor{cGrey}\textbf{1.0} &\cellcolor{cGrey}\textbf{11.9} &\cellcolor{cGrey}\textbf{28.9} &\cellcolor{cGrey}\textbf{48.2} &\cellcolor{cGrey}\textbf{57.6} &\cellcolor{cGrey}\textbf{6.0} &\cellcolor{cGrey}43.3  \\
    \hline 
    {\color{cGreen}\textit{\small{CLIP-ViT-B/16}}} & & & & & & & & & & \\
    X-Pool~\cite{gorti2022x} &48.2 &73.7 &82.6 &2.0 &12.7 &26.1 &46.8 &56.7 &7.0 &47.3 \\
    UATVR~\cite{fang2023uatvr} & 50.8 & 76.3 & 85.5 & 1.0 & 12.4  & -- &-- &-- & --&-- \\ 
    CLIP-ViP~\cite{xue2022clip} & \textbf{54.2} & \textbf{77.2} & 84.8 &1.0 & -- & 29.4 & 50.6 & 59.0 & 5.0 & -- \\
    \cellcolor{cGrey}T-MASS (\small{Ours})  &\cellcolor{cGrey}52.7 &\cellcolor{cGrey}77.1 &\cellcolor{cGrey}\textbf{85.6} &\cellcolor{cGrey} \textbf{1.0} &\cellcolor{cGrey}\textbf{10.5} &\cellcolor{cGrey}\textbf{30.3} &\cellcolor{cGrey}\textbf{52.2} &\cellcolor{cGrey}\textbf{61.3} &\cellcolor{cGrey}\textbf{5.0} &\cellcolor{cGrey}\textbf{40.1}  \\
    \hline
\end{tabular}}
\vspace{-5mm}
\end{center}
\caption{Text-to-video comparisons on MSRVTT~\cite{xu2016msr} and LSMDC~\cite{rohrbach2015dataset}. Bold denotes the best performance. ``--'': result is unavailable.
} 
\label{tab: benchmark-MSRVTT9K-LSMDC}
\vspace{-2mm}
\end{table*}

\begin{table*}[t]
\begin{center}
\scalebox{0.96}{
\begin{tabular}{c|ccccc|ccccc} 
    \hline 
     \multirow{2}{*}{Method} & \multicolumn{5}{c|}{DiDeMo Retrieval} & \multicolumn{5}{c}{VATEX Retrieval} \\
     \cline{2-11}
     & R@1 \small{$\uparrow$} & R@5 \small{$\uparrow$} & R@10 \small{$\uparrow$} & MdR \small{$\downarrow$} & MnR \small{$\downarrow$} & R@1 \small{$\uparrow$} & R@5 \small{$\uparrow$} & R@10 \small{$\uparrow$} & MdR  \small{$\downarrow$} & MnR \small{$\downarrow$} \\
    \hline
    {\color{cGreen}{\color{cGreen}\textit{\small{CLIP-ViT-B/32}}}} & & & & & & & & & & \\ 
    X-Pool~\cite{gorti2022x} & 44.6 & 73.2 & 82.0 & 2.0 & 15.4 &60.0 & 90.0 & 95.0 & 1.0 & 3.8   \\
    DiffusionRet~\cite{jin2023diffusionret}  & 46.7 & 74.7 & 82.7  & 2.0 & 14.3 & -- &-- & --&-- & --  \\
    UATVR~\cite{fang2023uatvr} & 43.1 & 71.8 & 82.3 & 2.0  & 15.1  & 61.3 & 91.0 & 95.6 & 1.0 & 3.3  \\
    CLIP-ViP~\cite{xue2022clip} & 48.6 & 77.1 & 84.4 & 2.0 & -- &-- &-- &-- &-- &--  \\
    \cellcolor{cGrey}T-MASS (\small{Ours})  &\cellcolor{cGrey}\textbf{50.9} &\cellcolor{cGrey}\textbf{77.2} &\cellcolor{cGrey}\textbf{85.3} &\cellcolor{cGrey}\textbf{1.0} &\cellcolor{cGrey} \textbf{12.1}&\cellcolor{cGrey}\textbf{63.0} &\cellcolor{cGrey} \textbf{92.3} &\cellcolor{cGrey}\textbf{96.4} &\cellcolor{cGrey}\textbf{1.0} &\cellcolor{cGrey}\textbf{3.2}  \\
    \hline 
    {\color{cGreen}\textit{\small{CLIP-ViT-B/16}}} & & & & & & & & & & \\
    X-Pool~\cite{gorti2022x} & 47.3 & 74.8 & 82.8 & 2.0 & 14.2 &62.6 &91.7 &96.0 &1.0 & 3.4 \\
    UATVR~\cite{fang2023uatvr} & 45.8 & 73.7 & 83.3 & 2.0 & 13.5 & 64.5 & 92.6 & 96.8 & 1.0 & 2.8  \\
    CLIP-ViP~\cite{xue2022clip} & 50.5 & 78.4  & 87.1 & 1.0 & -- & -- & --&-- &-- &--  \\
    \cellcolor{cGrey}T-MASS (\small{Ours})  &\cellcolor{cGrey}\textbf{53.3} &\cellcolor{cGrey}\textbf{80.1} &\cellcolor{cGrey}\textbf{87.7} &\cellcolor{cGrey}\textbf{1.0} &\cellcolor{cGrey}\textbf{9.8}  &\cellcolor{cGrey}\textbf{65.6} &\cellcolor{cGrey}\textbf{93.9} &\cellcolor{cGrey}\textbf{97.2} &\cellcolor{cGrey}\textbf{1.0}  &\cellcolor{cGrey}\textbf{2.7}  \\
    \hline
\end{tabular}}
\vspace{-5mm}
\end{center}
\caption{Text-to-video comparisons on DiDeMo~\cite{anne2017localizing} and VATEX~\cite{wang2019vatex}. Bold denotes the best performance. ``--'': result is unavailable. 
} 
\label{tab: benchmark-VATEX-DiDeMo}
\vspace{-4mm}
\end{table*}

\begin{table}[t]
\begin{center}
\scalebox{0.92}{
\begin{tabular}{c|ccccc} 
    \hline 
     Method & R@1 & R@5 & R@10 & MdR  & MnR \\
    \hline 
    {\color{cGreen}\textit{CLIP-ViT-B/32}} & & & & & \\
    CLIP4Clip~\cite{luo2022clip4clip} & 42.7 & 70.9 & 80.6 & 2.0 & 11.6   \\
    CenterCLIP~\cite{zhao2022centerclip}& 42.8 &71.7 &82.2& 2.0 &10.9 \\
    X-Pool~\cite{gorti2022x} & 44.4 & 73.3 & 84.0 & 2.0 & 9.0   \\
    TS2-Net~\cite{liu2022ts2} & 45.3 & 74.1 & 83.7 & 2.0 & 9.2 \\ 
    DiffusionRet~\cite{jin2023diffusionret} & 47.7 & 73.8 & 84.5 & 2.0 & 8.8  \\
    UATVR~\cite{fang2023uatvr} & 46.9 & 73.8 & 83.8 & 2.0 & 8.6 \\
    
    \cellcolor{cGrey}T-MASS (Ours)  &\cellcolor{cGrey}\textbf{47.7} &\cellcolor{cGrey}\textbf{78.0} &\cellcolor{cGrey}\textbf{86.3} &\cellcolor{cGrey}\textbf{2.0} &\cellcolor{cGrey}\textbf{8.0} \\ 
    \hline 
    {\color{cGreen}\textit{CLIP-ViT-B/16}} & & & & & \\
    X-Pool~\cite{gorti2022x} & 46.4 & 73.9  & 84.1 & 2.0  & 8.4   \\
    TS2-Net~\cite{liu2022ts2}  &46.6 & 75.9 & 84.9 & 2.0 & 8.9 \\ 
    CenterCLIP~\cite{zhao2022centerclip}& 47.7 &75.0& 83.3& 2.0 &10.2\\
    UATVR~\cite{fang2023uatvr} & 48.1& 76.3&85.4 &2.0 &8.0  \\
    
    \cellcolor{cGrey}T-MASS (Ours)  &\cellcolor{cGrey}\textbf{50.9} &\cellcolor{cGrey}\textbf{80.2} &\cellcolor{cGrey}\textbf{88.0} &\cellcolor{cGrey}\textbf{1.0} &\cellcolor{cGrey}\textbf{7.4} \\ 
    \hline 
\end{tabular}}
\vspace{-5mm}
\end{center}
\caption{Video-to-text comparisons on MSRVTT. } 
\label{tab: benchmark-MSRVTT9K-v2t}
\vspace{-6mm}
\end{table}

\begin{table}[t]
\begin{center}
\scalebox{0.94}{
\begin{tabular}{c|ccccc} 
    \hline 
     Method & R@1 & R@5 & R@10 & MdR  & MnR \\
    \hline 
    {\color{cGreen}{\color{cGreen}\textit{CLIP-ViT-B/32}}} & & & & & \\
    ClipBERT~\cite{lei2021less} &6.7  &17.3  &25.2  &32.0  &149.7\\
    CLIP4Clip~\cite{luo2022clip4clip} &9.9  &27.1  &36.8  &21.0  &85.4\\
    X-Pool~\cite{gorti2022x} &11.2  &28.3  &38.8  &20.0  &82.7\\
    \cellcolor{cGrey}T-MASS (Ours)  &\cellcolor{cGrey}\textbf{14.2} &\cellcolor{cGrey}\textbf{36.2}&\cellcolor{cGrey}\textbf{48.3} &\cellcolor{cGrey}\textbf{12.0} &\cellcolor{cGrey}\textbf{54.8} \\ 
    \hline 
    {\color{cGreen}\textit{CLIP-ViT-B/16}} & & & & & \\
     CLIP4Clip~\cite{luo2022clip4clip} &16.0 &38.2  &48.5  &12.0  &54.1\\
    X-Pool~\cite{gorti2022x} & 20.7 & 42.5 & 53.5 & 9.0 & 47.4 \\
    \cellcolor{cGrey}T-MASS (Ours)  &\cellcolor{cGrey}\textbf{26.7} &\cellcolor{cGrey}\textbf{51.7} &\cellcolor{cGrey}\textbf{63.9} &\cellcolor{cGrey}\textbf{5.0} &\cellcolor{cGrey}\textbf{30.0} \\ 
    \hline
\end{tabular}}
\vspace{-5mm}
\end{center}
\caption{Text-to-video comparisons on Charades~\cite{sigurdsson2016hollywood}.  } 
\label{tab: benchmark-Charades}
\vspace{-7mm}
\end{table}

\noindent\textbf{Datasets and Metrics}.
We adopt five benchmark datasets for the evaluation, including (1) \textbf{MSRVTT}~\cite{xu2016msr} that contains $10$K video clips, where each has $20$ captions. We follow the 1K-A testing split~\cite{Liu2019UseWY}. (2) \textbf{LSMDC}~\cite{rohrbach2015dataset} incorporating $118081$ clips from $202$ movies, where each one is paired with a text description. Following~\cite{gabeur2020multi,gorti2022x}, we adopt the testing data with $1000$ videos.  (3) \textbf{DiDeMo}~\cite{anne2017localizing} consists of $10642$ clips and $40543$ captions in total. We use the training/testing data following~\cite{luo2022clip4clip,jin2023diffusionret}. (4) \textbf{Charades}~\cite{sigurdsson2016hollywood} contains $9848$ video clips, where each corresponds to a text description. We adopt the same split protocol as in~\cite{lin2022eclipse}. (5) \textbf{VATEX}~\cite{wang2019vatex} consists of  $34,991$ video clips, where each corresponds to multiple text descriptions. We follow the train-test split of \cite{chen2020fine}. Recall at rank $\{1,5,10\}$ (R@1, R@5, and R@10), Median Rank (MdR), and Mean Rank (MnR) are adopted to evaluate the retrieval performance. 

\textbf{Implementation Details}.
We employ X-Pool~\cite{gorti2022x} as baseline. Both backbone models of CLIP~\cite{radford2021learning} (both ViT-B/32 and ViT-B/16) are leveraged for the feature extraction, following previous methods~\cite{xue2022clip,gorti2022x}. We keep the configurations the same as X-Pool, such that setting dimension $d=512$, weight decay as $0.2$, and dropout as $0.3$. For the training, we set the batch size as $32$ for both backbones and different datasets. We keep the same initial learning rate of $1e-5$ to train the feature fusion module $\psi(\cdot)$ and the proposed similarity-aware radius module $\mathcal{R}$ ($3e-5$ for MSRVTT). The CLIP model is fine-tuned with a learning rate of $1e-6$. We train the models for $5$ epochs with the AdamW~\cite{loshchilov2017decoupled} optimizer. Following CLIP, we employ a cosine schedule~\cite{loshchilov2016sgdr} with a warm-up proportion of $0.1$. We uniformly sample $12$ frames from the video clips upon different datasets. All the frames are resized to $224\times224$. We perform experiments on an A6000 GPU. We set sampling trials $T'=20$ during inference. Some methods use larger batch size and larger frames numbers for different datasets. We keep it consistent for our method by using  batch size as $32$ and frame number as $12$ for all datasets.  More results and discussions are provided in supplementary.

\subsection{Performance Comparison}
\label{subsec: performance}
We compare the text-to-video retrieval performance of T-MASS with previous methods on five benchmark datasets.  We find that T-MASS not only improves the baseline X-Pool by a large margin on all metrics, but also achieves state-of-the-art performance compared with most recent methods. As shown in Table~\ref{tab: benchmark-MSRVTT9K-LSMDC}, T-MASS improves improves CLIP-ViP $3.3\%$ at R@1 on LSMDC ViT-B/32 model. In Table~\ref{tab: benchmark-VATEX-DiDeMo}, T-MASS improves X-Pool by $6.0\%$ at R@1 on DiDeMo upon ViT-b/16. By observation, the proposed method shows a consistent performance boost on versatile datasets and different scales of model size. 
There exists one scenario under MSRVTT and ViT-B/16 that CLIP-ViT works better than T-MASS. Note that besides the retrieval data, CLIP-ViP also adopts additional datasets, \emph{e.g.}, WebVid-2.5M~\cite{bain2021frozen} and HD-VILA-100M~\cite{xue2022advancing} to further empower the post-pretraining, potentially better adapt the CLIP. Employing more data especially benefits the larger model of ViT-B/16. T-MASS outperforms CLIP-ViP on other datasets and backbones. To save the computational cost, this work does not include the additional multi-modal data. As shown Table~\ref{tab: benchmark-MSRVTT9K-v2t},  T-MASS also enables the best performance for video-to-text retrieval. CLIP-ViP is skipped as the result is unavailable.  
In summary, since T-MASS empowers text embedding with more flexibility, it potentially explores more possibilities in text-video alignment. We provide a more in-depth analysis in the following. 

\begin{table*}[t]
\begin{center}
\scalebox{0.96}{
\begin{tabular}{c|ccccc|ccccc} 
    \hline 
     \multirow{2}{*}{Radius $\mathcal{R}$} & \multicolumn{5}{c|}{MSRVTT Retrieval} & \multicolumn{5}{c}{DiDeMo Retrieval} \\
     \cline{2-11}
     & R@1 $\uparrow$ & R@5 $\uparrow$ & R@10 $\uparrow$ & MdR $\downarrow$ & MnR $\downarrow$ & R@1 $\uparrow$ & R@5 $\uparrow$ & R@10 $\uparrow$ & MdR  $\downarrow$ & MnR $\downarrow$\\
    \hline  
     w/o $\mathcal{R}$ & 46.9 & 72.8 & 82.2 & 2.0 & 14.3 & 44.6 & 73.2 & 82.0 & 2.0 & 15.4  \\

    \cellcolor{cGrey}$\text{exp}(\frac{1}{T'}\sum\nolimits{S_i})$ &\cellcolor{cGrey} 48.7&\cellcolor{cGrey}74.7&\cellcolor{cGrey}83.7&\cellcolor{cGrey}2.0&\cellcolor{cGrey}12.7&\cellcolor{cGrey}48.0&\cellcolor{cGrey}75.4&\cellcolor{cGrey}85.0&\cellcolor{cGrey}2.0&\cellcolor{cGrey}13.0\\
    \cellcolor{cGrey}$\text{exp}(\frac{\theta}{T'}\sum\nolimits{S_i})$ &\cellcolor{cGrey}\textbf{49.2}&\cellcolor{cGrey}75.7&\cellcolor{cGrey}84.7&\cellcolor{cGrey}2.0&\cellcolor{cGrey}\textbf{11.7}&\cellcolor{cGrey}49.7&\cellcolor{cGrey}75.8&\cellcolor{cGrey}85.3&\cellcolor{cGrey}2.0&\cellcolor{cGrey}12.6\\
    $\cellcolor{cGrey}\text{exp}(\mathbf{S}\mathbf{W})$ &\cellcolor{cGrey} 49.1&\cellcolor{cGrey}\textbf{75.7}&\cellcolor{cGrey}\textbf{85.7}&\cellcolor{cGrey}\textbf{2.0}&\cellcolor{cGrey}11.9&\cellcolor{cGrey}\textbf{49.8}&\cellcolor{cGrey}\textbf{78.1}&\cellcolor{cGrey}\textbf{86.0}&\cellcolor{cGrey}\textbf{2.0}&\cellcolor{cGrey}\textbf{11.8} \\
    \hline
\end{tabular}}
\vspace{-5mm}
\end{center}
\caption{Model discussion on similarity-aware radius module design. We perform experiments on MSRVTT and DiDeMo. CLIP-ViT-B/32 is adopted. Notebaly, ``w/o $\mathcal{R}$'' denotes the baseline of X-Pool~\cite{gorti2022x}. We choose $\text{exp}(\mathbf{S}\mathbf{W})$ for the final performance comparison.
} 
\label{tab: R ablation}
\vspace{-2mm}
\end{table*}

\begin{table*}[t] 
        \label{tab: discussion}
	\subfloat[Ablation study of losses and text embedding on MSRVTT~\cite{xu2016msr}. \label{tab: ablation-loss}]{
		\scalebox{0.96}{
            \begin{tabular}{cccc|ccccc} 
    \hline 
    $\mathbf{t}_s$ & $\mathcal{L}_\text{ce}$ & $ \mathcal{L}_s$ & $\mathcal{L}_\text{sup}$ & R@1 & R@5 & R@10 & MdR  & MnR \\
    \hline 
    \ding{55} & \ding{51} & \ding{55}&   \ding{55} & 46.9 & 72.8 & 82.2 & 2.0 & 14.3 \\
    \ding{51} & \ding{51} & \ding{51} & \ding{55}   & 48.5 &74.8 &84.3 &2.0 & 12.3\\
    \ding{51} & \ding{55} & \ding{51}&  \ding{55} &49.1 &\textbf{75.7} &\textbf{85.7} &2.0 &11.9 \\
    \cellcolor{cGrey}\ding{51} & \cellcolor{cGrey}\ding{55}  & \cellcolor{cGrey}\ding{51} & \cellcolor{cGrey}\ding{51}   &\cellcolor{cGrey}\textbf{50.2} &\cellcolor{cGrey}75.3 &\cellcolor{cGrey}85.1 &\cellcolor{cGrey}\textbf{1.0} &\cellcolor{cGrey}\textbf{11.9} \\
    \hline 
    \end{tabular}
    \vspace{-0.4cm}}} 
	\subfloat[ Discussion of stochastic sampling trails on MSRVTT-1K. \label{tab: sampling}]{ 
		\scalebox{0.96}{
            \begin{tabular}{c|ccccc} 
            \hline
        \#Trials ($M$) & R@1 & R@5 & R@10 &  MdR &  MnR \\
            \hline
            w/o sampling & 44.4 & 72.4 & 81.9 & 2.0 & 13.1 \\
            \cellcolor{cGrey}5 &\cellcolor{cGrey} 46.8&\cellcolor{cGrey}74.7 &\cellcolor{cGrey}84.0 &\cellcolor{cGrey}2.0 &\cellcolor{cGrey}12.5 \\
            \cellcolor{cGrey}10 &\cellcolor{cGrey}50.0 &\cellcolor{cGrey}75.2 &\cellcolor{cGrey}84.1 &\cellcolor{cGrey}2.0 &\cellcolor{cGrey}12.3 \\
            \cellcolor{cGrey}20 &\cellcolor{cGrey}\textbf{50.2} &\cellcolor{cGrey}\textbf{75.3} &\cellcolor{cGrey}\textbf{85.1} &\cellcolor{cGrey}\textbf{1.0} &\cellcolor{cGrey}\textbf{11.9} \\ 
            \hline
            \end{tabular}
            }}
            \vspace{-1.5mm}
            \caption{Discussion of the text representation, learning objectives, and number of trials. $\mathbf{t}_s$ denotes stochastic embedding, relative to text embedding $\mathbf{t}$. $\mathcal{L}_\text{ce}$ denotes  symmetric cross entropy loss upon Eq.~\eqref{eq: contrastive loss}. $\mathcal{L}_s$ computes upon $\mathbf{t}_s$ and $\mathbf{v}$. $\mathcal{L}_\text{sup}$ denotes support text  regularization.  }	 
\vspace{-5mm}
\end{table*}

\begin{figure*}[t]
\scriptsize
\hspace{-1mm}
\resizebox{0.96\textwidth}{!}{
\begin{tabular}{cccc}
\begin{adjustbox}{valign=t}
    \begin{tabular}{c}
    \includegraphics[width=0.64\textwidth,angle=0]{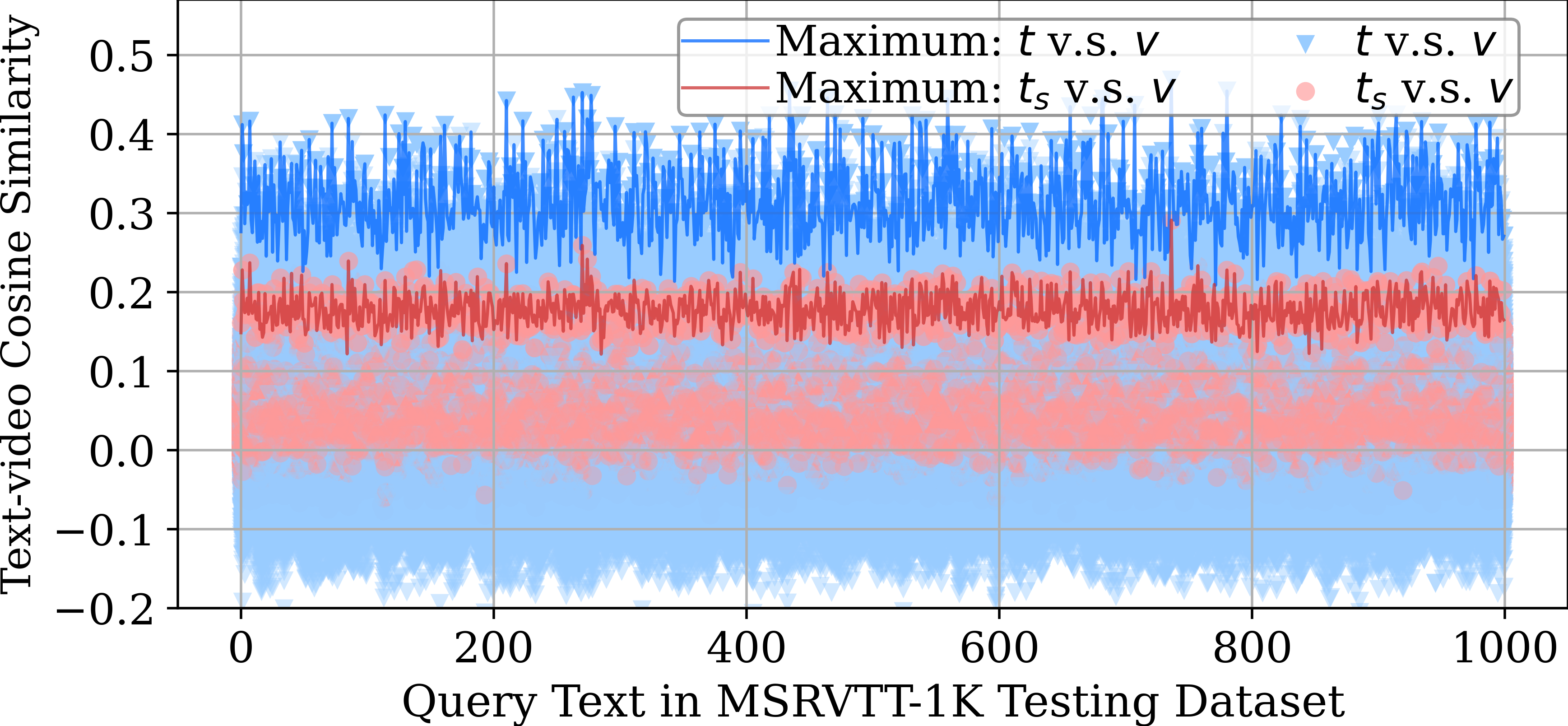}
    \end{tabular}
\end{adjustbox}
\hspace{-2mm} 
\begin{adjustbox}{valign=t}
    \begin{tabular}{c}
    \includegraphics[width=0.35\textwidth,angle=0]{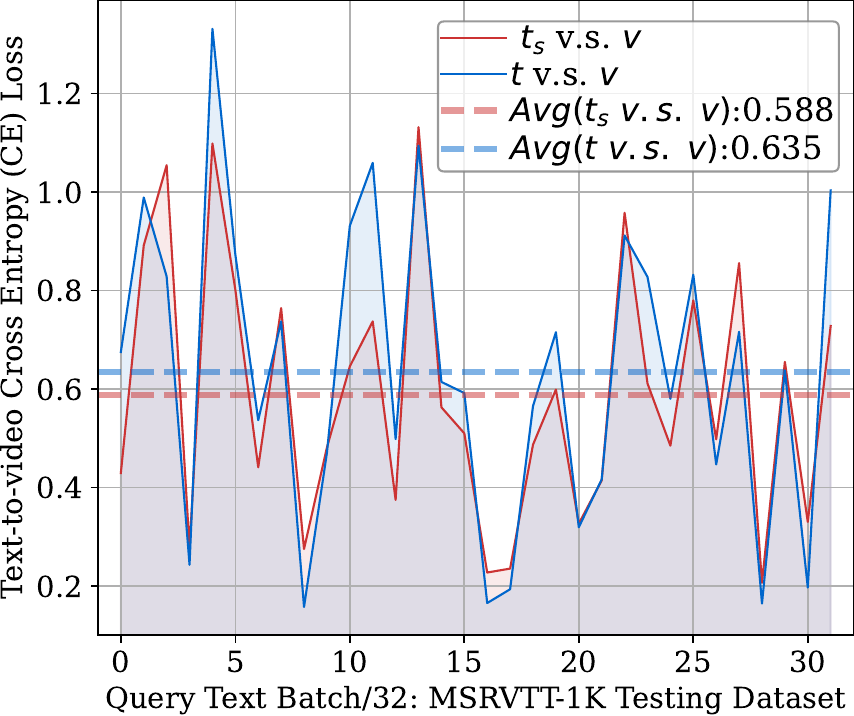}
    \end{tabular}
\end{adjustbox}
\end{tabular}
}
\vspace{-2.4mm}
\caption{
Analysis of stochastic text embedding $\mathbf{t}_s$, text embedding $\mathbf{t}$, and video embedding $\mathbf{v}$ in a joint space. \textbf{\textit{Left}}: Cosine similarities of \textbf{irrelevant} text-video pairs in embedding space. \textbf{\textit{Right}}: Cross entropy values of \textbf{relevant} text-video pairs in embedding space. The proposed stochastic text embedding allows a lower similarity for irrelevant pairs and enables lower cross entropy loss for relevant pairs. 
}
\label{fig: postive-negative}
\vspace{-5mm}
\end{figure*}

\subsection{Model Discussion}
\label{subsec: discussion}

\textbf{Similarity-Aware Radius}.
In Table~\ref{tab: R ablation}, we provide three options to implement the similarity-aware radius module, as introduced in Section~\ref{subsec: T-MASS}. Specifically,  $\text{exp}(\frac{1}{T'}\sum\nolimits{S_i})$ denotes only using cosine similarity to implement the radius, which is unlearnable. We further incorporate a learnable scalar $\theta$, resulting $\text{exp}(\frac{\theta}{T'}\sum\nolimits{S_i})$, or using a linear layer, yielding $\text{exp}(\mathbf{S}\mathbf{W})$. 
Note that ``w/o $\mathcal{R}$'' denotes the baseline of X-Pool~\cite{gorti2022x}.
The design of the $\mathcal{R}$ brings a clear performance boost compared with the baseline, \emph{i.e.}, $>1.5\%+$ at R@1 on MSRVTT and $>3\%+$ at R@1 on DiDeMo. 
This indicates that representing text as a semantics range can indeed further benefit the retrieval on these two datasets, compared with $\mathbf{t}$. 
Besides, using a learnable module to can further boost the performance as the expressiveness and flexibility of the mass improve. The design of  $\text{exp}(\mathbf{S}\mathbf{W})$ works best in most cases (especially on DiDeMo), owning to a stronger modeling capacity. Interestingly, only using a learnable scalar also enables strong performance, indicating that our approach is not sensitive to the network design of $\mathcal{R}$. We adopt $\text{exp}(\mathbf{S}\mathbf{W})$ in our final model.

\textbf{Ablation Study}.
We provide an ablation study on MSRVTT in terms of text representation and learning objectives in Table~\ref{tab: ablation-loss}.  Firstly, we show the baseline of X-Pool (top row). Based on X-Pool, we substitute text embedding $\mathbf{t}$ with $\mathbf{t}_s$ and correspondingly add a $\mathcal{L}_s$, obtaining $1.6\%$ boost at R@1. This shows the superiority of $\textbf{t}_s$ over $\mathbf{t}$.  We further evaluate the effect of the original loss $\mathcal{L}_\text{ce}$ under the regime of the stochastic embedding $\mathbf{t}_s$. 
By comparison, highlighting the text embedding $\mathbf{t}$ with $\mathcal{L}_\text{ce}$ undermines the performance (2nd and 3rd rows of Table~\ref{tab: ablation-loss}). This is because further regularizing $\mathbf{t}$ can lead to a biased text mass learning, misleading the retrieval. Rather, we adopt a support text vector $\mathbf{t}_\text{sup}$ as a proxy to control the scale and shift of the text mass.  Since $\mathbf{t}_\text{sup}$ locates at the surface of the text mass upon Eq.~\eqref{eq: support text}, as shown in Fig.~\ref{fig: support loss}, controlling support text embedding can affect the whole text mass. We adopt the last setting (the 4th row in Table~\ref{tab: ablation-loss}) in our final model. 

\textbf{Inference Discussion}.
We discuss the number of sampling trials for inference in Table~\ref{tab: sampling}. For ``w/o sampling'', we still use the original $\mathbf{t}$ during the metric computation. This gives a sub-optimal performance as there is no exploitation of the text mass. As the number of trails $M$ increases from $5$ to $20$, the proposed method enables better performance by exploiting more possibilities of the stochastic embedding $\mathbf{t}_s$ corresponding to the semantics of the raw text $t$. The number of $5$ may not be enough to explore the text mass, leading to a sub-optimal result.
Note that the performance tends to be stable from $M=10$ to $M=20$. We choose $20$ in our final model, considering the trade-off between the performance and computational cost.

\textbf{Further Analysis on T-MASS}.
We further analyze the behavior of the T-MASS by observing the stochastic text embedding $\mathbf{t}_s$, text embedding $\mathbf{t}$, and video embedding $\mathbf{v}$ in the same joint space.  As shown in Fig.~\ref{fig: postive-negative} \textit{left}, we collect the text-video cosine similarity values for all irrelevant pairs in MSRVTT-1K, comparing between $\{\mathbf{t}^i~ \text{v.s.}~ \mathbf{v}^j\}$ and $ \{\mathbf{t}_s^i ~\text{v.s.}~ \mathbf{v}^j\}$, where $i \neq j$. For each query text, we plot similarity values to all irrelevant videos (\emph{i.e.}, $999$) and highlight the maximum value (red and blue curves). The smaller similarity values are, the better irrelevant text-video pairs are aligned and thus potentially benefits the retrieval. By observation, using stochastic embedding $\mathbf{t}_s$ gives a better result than $\mathbf{t}$ (red curve is lower).
This indicates that the irrelevant $\mathbf{t}$ and $\mathbf{v}$ can be close to each other, which may impose more risks for mismatching. We also visualize the cross-entropy loss values of relevant text-video pairs in Fig.~\ref{fig: postive-negative} \textit{right}. By average, the proposed $\mathbf{t}_s$ enables lower entropy, which reflects  higher similarities for relevant $t$-$v$ pairs, ensuring a more promising and accurate retrieval. In summary, both comparisons (Fig.~\ref{fig: postive-negative} \textit{left} and \textit{right}) show that T-MASS indeed enables a better text-video embedding alignment.

\begin{figure}[t] 
\centering 
\includegraphics[width=.475\textwidth]{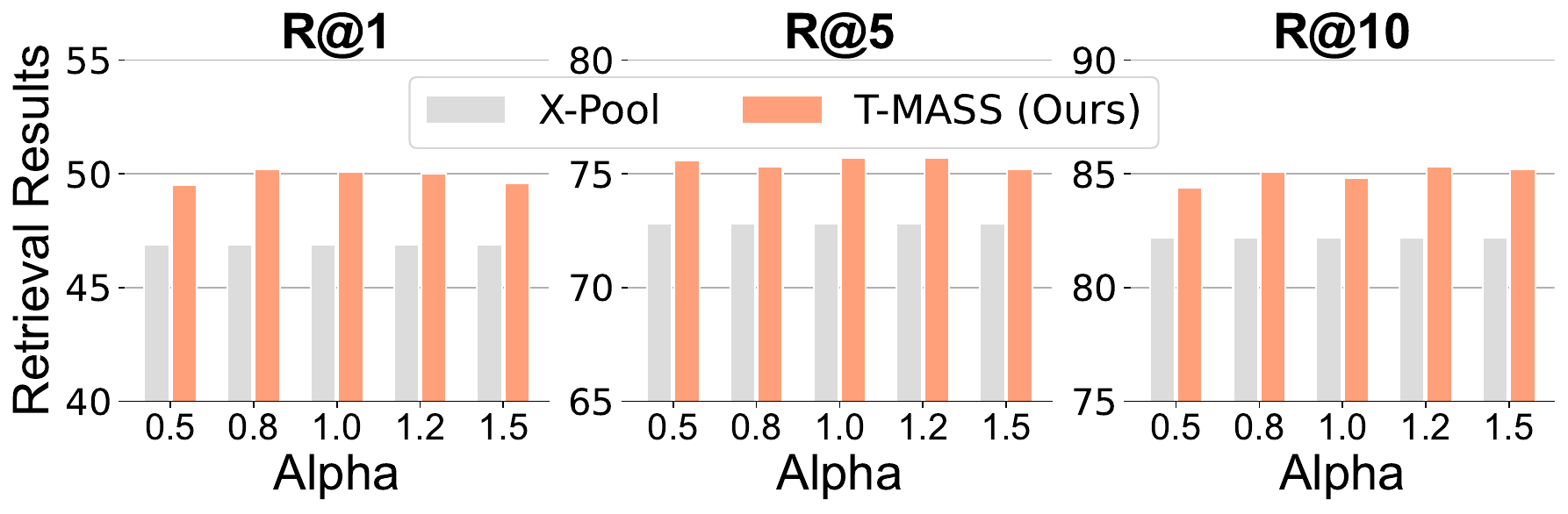}
\vspace{-7mm}
\caption{Discussion of support text regularization weight $\alpha$. We compare T-MASS with baseline X-Pool~\cite{gorti2022x} on MSRVTT~\cite{xu2016msr}.  } 
\label{fig: Support weight discuss}
\vspace{-5.mm} 
\end{figure}

\begin{figure}[t] 
\centering 
\includegraphics[width=.475\textwidth]{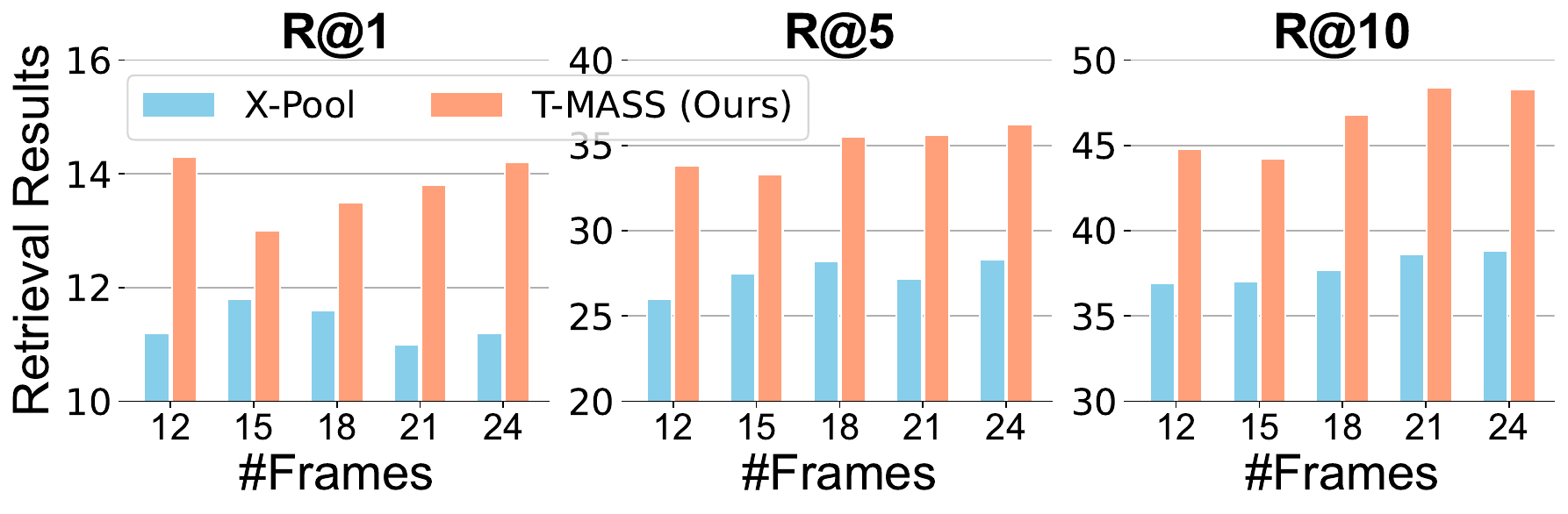}
\vspace{-7mm}
\caption{Performance boost under different \#frames ($T'$). We compare T-MASS with the baseline X-Pool~\cite{gorti2022x} on Charades~\cite{sigurdsson2016hollywood}.  } 
\label{fig: frame discuss}
\vspace{-4.9mm} 
\end{figure}

\textbf{Hyperparameter Discussion}. 
Fig.~\ref{fig: Support weight discuss} discusses the effect of the support text regularization. Retrieval performance under different penalties, such as $\alpha=\{0.5, 0.8, 1.0, 1.2, 1.5\}$ are presented. The performance of X-Pool is provided as a reference. We achieve the best performance at $\alpha=1.2$. 
We also discuss the effect of the \#frames on Charades in Fig.~\ref{fig: frame discuss}. Specifically, we report performance with $T'=\{12, 15, 18, 21, 24\}$. T-MASS enables a notable performance boost under different $T'$ values. Owning to the text mass learning, this method demonstrates robustness to different configurations of input videos.

\section{Conclusions}
\label{sec:conclusion}

This work studied the text-video retrieval task by drawing attention to the text side --  text is hard to fully describe the semantics of a video, implying that text embedding may not be expressive enough to capture or align to the video -- based on which we opted to enrich the text embedding with more flexibility and resilience. We introduced T-MASS, where text is modeled as a stochastic embedding, facilitating joint learning of the text mass and video points. Our method incorporated similarity-aware radius modeling and a support text vector as regularization to better align relevant/irrelevant text-video pairs and encourage text semantics determination. Experiments on five datasets demonstrated that T-MASS achieved state-of-the-art performance. We hope this work will inspire future endeavors from the perspective of text in advancing text-video retrieval.

\section{Acknowledgement}
\label{sec:ack}

This research was partially supported by the DEVCOM Army Research Laboratory under Contract W911QX-21-D-0001.
\clearpage
{
    \small
    \bibliographystyle{ieeenat_fullname}
    \bibliography{main}

\begin{thebibliography}{61}
\providecommand{\natexlab}[1]{#1}
\providecommand{\url}[1]{\texttt{#1}}
\expandafter\ifx\csname urlstyle\endcsname\relax
  \providecommand{\doi}[1]{doi: #1}\else
  \providecommand{\doi}{doi: \begingroup \urlstyle{rm}\Url}\fi

\bibitem[Akbari et~al.(2021)Akbari, Yuan, Qian, Chuang, Chang, Cui, and Gong]{NEURIPS2021_cb3213ad}
Hassan Akbari, Liangzhe Yuan, Rui Qian, Wei-Hong Chuang, Shih-Fu Chang, Yin Cui, and Boqing Gong.
\newblock Vatt: Transformers for multimodal self-supervised learning from raw video, audio and text.
\newblock In \emph{NeurIPS}, 2021.

\bibitem[Anne~Hendricks et~al.(2017)Anne~Hendricks, Wang, Shechtman, Sivic, Darrell, and Russell]{anne2017localizing}
Lisa Anne~Hendricks, Oliver Wang, Eli Shechtman, Josef Sivic, Trevor Darrell, and Bryan Russell.
\newblock Localizing moments in video with natural language.
\newblock In \emph{ICCV}, 2017.

\bibitem[Bain et~al.(2021)Bain, Nagrani, Varol, and Zisserman]{bain2021frozen}
Max Bain, Arsha Nagrani, G{\"u}l Varol, and Andrew Zisserman.
\newblock Frozen in time: A joint video and image encoder for end-to-end retrieval.
\newblock In \emph{ICCV}, 2021.

\bibitem[Bain et~al.(2022)Bain, Nagrani, Varol, and Zisserman]{bain2022clip}
Max Bain, Arsha Nagrani, G{\"u}l Varol, and Andrew Zisserman.
\newblock A clip-hitchhiker's guide to long video retrieval.
\newblock \emph{arXiv}, 2022.

\bibitem[Bogolin et~al.(2022)Bogolin, Croitoru, Jin, Liu, and Albanie]{bogolin2022cross}
Simion-Vlad Bogolin, Ioana Croitoru, Hailin Jin, Yang Liu, and Samuel Albanie.
\newblock Cross modal retrieval with querybank normalisation.
\newblock In \emph{CVPR}, 2022.

\bibitem[Chen et~al.(2020)Chen, Zhao, Jin, and Wu]{chen2020fine}
Shizhe Chen, Yida Zhao, Qin Jin, and Qi Wu.
\newblock Fine-grained video-text retrieval with hierarchical graph reasoning.
\newblock In \emph{CVPR}, 2020.

\bibitem[Cheng et~al.(2021)Cheng, Lin, Wu, Yang, and Shen]{cheng2021improving}
Xing Cheng, Hezheng Lin, Xiangyu Wu, Fan Yang, and Dong Shen.
\newblock Improving video-text retrieval by multi-stream corpus alignment and dual softmax loss.
\newblock \emph{arXiv}, 2021.

\bibitem[Croitoru et~al.(2021)Croitoru, Bogolin, Leordeanu, Jin, Zisserman, Albanie, and Liu]{croitoru2021teachtext}
Ioana Croitoru, Simion-Vlad Bogolin, Marius Leordeanu, Hailin Jin, Andrew Zisserman, Samuel Albanie, and Yang Liu.
\newblock Teachtext: Crossmodal generalized distillation for text-video retrieval.
\newblock In \emph{CVPR}, 2021.

\bibitem[Deng et~al.(2023)Deng, Chen, Qin, Chen, and Wu]{deng2023prompt}
Chaorui Deng, Qi Chen, Pengda Qin, Da Chen, and Qi Wu.
\newblock Prompt switch: Efficient clip adaptation for text-video retrieval.
\newblock In \emph{ICCV}, 2023.

\bibitem[Dong et~al.(2022)Dong, Chen, Zhang, Yang, Chen, Li, and Wang]{PRVR}
Jianfeng Dong, Xianke Chen, Minsong Zhang, Xun Yang, Shujie Chen, Xirong Li, and Xun Wang.
\newblock Partially relevant video retrieval.
\newblock In \emph{ACM MM}, 2022.

\bibitem[Dosovitskiy et~al.(2020)Dosovitskiy, Beyer, Kolesnikov, Weissenborn, Zhai, Unterthiner, Dehghani, Minderer, Heigold, Gelly, et~al.]{dosovitskiy2020image}
Alexey Dosovitskiy, Lucas Beyer, Alexander Kolesnikov, Dirk Weissenborn, Xiaohua Zhai, Thomas Unterthiner, Mostafa Dehghani, Matthias Minderer, Georg Heigold, Sylvain Gelly, et~al.
\newblock An image is worth 16x16 words: Transformers for image recognition at scale.
\newblock \emph{arXiv}, 2020.

\bibitem[Dzabraev et~al.(2021)Dzabraev, Kalashnikov, Komkov, and Petiushko]{dzabraev2021mdmmt}
Maksim Dzabraev, Maksim Kalashnikov, Stepan Komkov, and Aleksandr Petiushko.
\newblock Mdmmt: Multidomain multimodal transformer for video retrieval.
\newblock In \emph{CVPR}, 2021.

\bibitem[Fang et~al.(2023)Fang, Liu, Zhou, Yang, Song, Li, Wang, Ji, Ouyang, et~al.]{fang2023uatvr}
Bo Fang, Chang Liu, Yu Zhou, Min Yang, Yuxin Song, Fu Li, Weiping Wang, Xiangyang Ji, Wanli Ouyang, et~al.
\newblock Uatvr: Uncertainty-adaptive text-video retrieval.
\newblock In \emph{ICCV}, 2023.

\bibitem[Fang et~al.(2021)Fang, Xiong, Xu, and Chen]{fang2021clip2video}
Han Fang, Pengfei Xiong, Luhui Xu, and Yu Chen.
\newblock Clip2video: Mastering video-text retrieval via image clip.
\newblock \emph{arXiv preprint arXiv:2106.11097}, 2021.

\bibitem[Gabeur et~al.(2020)Gabeur, Sun, Alahari, and Schmid]{gabeur2020multi}
Valentin Gabeur, Chen Sun, Karteek Alahari, and Cordelia Schmid.
\newblock Multi-modal transformer for video retrieval.
\newblock In \emph{ECCV}, 2020.

\bibitem[Gao et~al.(2021)Gao, Liu, Chen, Chang, Zhang, and Yuan]{CLIP2TV}
Zijian Gao, Jingyu Liu, Sheng Chen, Dedan Chang, Hao Zhang, and Jinwei Yuan.
\newblock {CLIP2TV:} an empirical study on transformer-based methods for video-text retrieval.
\newblock \emph{CoRR}, 2021.

\bibitem[Gorti et~al.(2022)Gorti, Vouitsis, Ma, Golestan, Volkovs, Garg, and Yu]{gorti2022x}
Satya~Krishna Gorti, No{\"e}l Vouitsis, Junwei Ma, Keyvan Golestan, Maksims Volkovs, Animesh Garg, and Guangwei Yu.
\newblock X-pool: Cross-modal language-video attention for text-video retrieval.
\newblock In \emph{CVPR}, 2022.

\bibitem[Guan et~al.(2023)Guan, Pei, Shao, Liu, Li, Gu, Xu, Xu, Yan, and Lam]{guan2023pidro}
Peiyan Guan, Renjing Pei, Bin Shao, Jianzhuang Liu, Weimian Li, Jiaxi Gu, Hang Xu, Songcen Xu, Youliang Yan, and Edmund~Y Lam.
\newblock Pidro: Parallel isomeric attention with dynamic routing for text-video retrieval.
\newblock In \emph{ICCV}, 2023.

\bibitem[Han et~al.(2022)Han, Chen, Zhang, Wang, and Chen]{ADVMM}
Ning Han, Jingjing Chen, Hao Zhang, Huanwen Wang, and Hao Chen.
\newblock Adversarial multi-grained embedding network for cross-modal text-video retrieval.
\newblock \emph{ACM MM}, 2022.

\bibitem[Huang et~al.(2023)Huang, Li, Feng, Wu, Sun, and Ji]{10205280}
J. Huang, Y. Li, J. Feng, X. Wu, X. Sun, and R. Ji.
\newblock Clover: Towards a unified video-language alignment and fusion model.
\newblock In \emph{CVPR}, 2023.

\bibitem[Ibrahimi et~al.(2023)Ibrahimi, Sun, Wang, Garg, Sanan, and Omar]{ibrahimi2023audio}
Sarah Ibrahimi, Xiaohang Sun, Pichao Wang, Amanmeet Garg, Ashutosh Sanan, and Mohamed Omar.
\newblock Audio-enhanced text-to-video retrieval using text-conditioned feature alignment.
\newblock In \emph{ICCV}, 2023.

\bibitem[Ji et~al.(2023)Ji, Tu, Jiang, Kong, Cai, Zhao, Wang, Yang, and Liu]{10204414}
Y. Ji, R. Tu, J. Jiang, W. Kong, C. Cai, W. Zhao, H. Wang, Y. Yang, and W. Liu.
\newblock Seeing what you miss: Vision-language pre-training with semantic completion learning.
\newblock In \emph{CVPR}, 2023.

\bibitem[Jin et~al.(2022)Jin, Huang, Liu, Wu, Ge, Song, Clifton, and Chen]{jin2022expectation}
Peng Jin, Jinfa Huang, Fenglin Liu, Xian Wu, Shen Ge, Guoli Song, David Clifton, and Jie Chen.
\newblock Expectation-maximization contrastive learning for compact video-and-language representations.
\newblock In \emph{NeurIPS}, 2022.

\bibitem[Jin et~al.(2023{\natexlab{a}})Jin, Huang, Xiong, Tian, Liu, Ji, Yuan, and Chen]{jin2023video}
Peng Jin, Jinfa Huang, Pengfei Xiong, Shangxuan Tian, Chang Liu, Xiangyang Ji, Li Yuan, and Jie Chen.
\newblock Video-text as game players: Hierarchical banzhaf interaction for cross-modal representation learning.
\newblock In \emph{CVPR}, 2023{\natexlab{a}}.

\bibitem[Jin et~al.(2023{\natexlab{b}})Jin, Li, Cheng, Huang, Wang, Yuan, Liu, and Chen]{jin2023text}
Peng Jin, Hao Li, Zesen Cheng, Jinfa Huang, Zhennan Wang, Li Yuan, Chang Liu, and Jie Chen.
\newblock Text-video retrieval with disentangled conceptualization and set-to-set alignment.
\newblock In \emph{IJCAI}, 2023{\natexlab{b}}.

\bibitem[Jin et~al.(2023{\natexlab{c}})Jin, Li, Cheng, Li, Ji, Liu, Yuan, and Chen]{jin2023diffusionret}
Peng Jin, Hao Li, Zesen Cheng, Kehan Li, Xiangyang Ji, Chang Liu, Li Yuan, and Jie Chen.
\newblock Diffusionret: Generative text-video retrieval with diffusion model.
\newblock In \emph{ICCV}, 2023{\natexlab{c}}.

\bibitem[Kingma and Welling(2013)]{kingma2013auto}
Diederik~P Kingma and Max Welling.
\newblock Auto-encoding variational bayes.
\newblock \emph{arXiv preprint arXiv:1312.6114}, 2013.

\bibitem[Lei et~al.(2021)Lei, Li, Zhou, Gan, Berg, Bansal, and Liu]{lei2021less}
Jie Lei, Linjie Li, Luowei Zhou, Zhe Gan, Tamara~L Berg, Mohit Bansal, and Jingjing Liu.
\newblock Less is more: Clipbert for video-and-language learning via sparse sampling.
\newblock In \emph{CVPR}, 2021.

\bibitem[Li et~al.(2023{\natexlab{a}})Li, Gan, Lin, Lin, Liu, Liu, and Wang]{Li_2023_CVPR}
Linjie Li, Zhe Gan, Kevin Lin, Chung-Ching Lin, Zicheng Liu, Ce Liu, and Lijuan Wang.
\newblock Lavender: Unifying video-language understanding as masked language modeling.
\newblock In \emph{CVPR}, 2023{\natexlab{a}}.

\bibitem[Li et~al.(2023{\natexlab{b}})Li, Xie, Zhao, Xie, Ge, Zheng, Zhao, and Zhang]{li2023progressive}
Pandeng Li, Chen-Wei Xie, Liming Zhao, Hongtao Xie, Jiannan Ge, Yun Zheng, Deli Zhao, and Yongdong Zhang.
\newblock Progressive spatio-temporal prototype matching for text-video retrieval.
\newblock In \emph{ICCV}, 2023{\natexlab{b}}.

\bibitem[Li et~al.(2023{\natexlab{c}})Li, Min, Tripathi, and Vasconcelos]{10205049}
Y. Li, K. Min, S. Tripathi, and N. Vasconcelos.
\newblock Svitt: Temporal learning of sparse video-text transformers.
\newblock In \emph{CVPR}, 2023{\natexlab{c}}.

\bibitem[Lin et~al.(2022{\natexlab{a}})Lin, Wu, Liang, Zhang, Ge, Zheng, and Shen]{lin2022text}
Chengzhi Lin, Ancong Wu, Junwei Liang, Jun Zhang, Wenhang Ge, Wei-Shi Zheng, and Chunhua Shen.
\newblock Text-adaptive multiple visual prototype matching for video-text retrieval.
\newblock \emph{NeurIPS}, 2022{\natexlab{a}}.

\bibitem[Lin et~al.(2022{\natexlab{b}})Lin, Lei, Bansal, and Bertasius]{lin2022eclipse}
Yan-Bo Lin, Jie Lei, Mohit Bansal, and Gedas Bertasius.
\newblock Eclipse: Efficient long-range video retrieval using sight and sound.
\newblock In \emph{ECCV}, 2022{\natexlab{b}}.

\bibitem[Liu et~al.(2019)Liu, Albanie, Nagrani, and Zisserman]{Liu2019UseWY}
Yang Liu, Samuel Albanie, Arsha Nagrani, and Andrew Zisserman.
\newblock Use what you have: Video retrieval using representations from collaborative experts.
\newblock In \emph{BMVC}, 2019.

\bibitem[Liu et~al.(2022{\natexlab{a}})Liu, Chen, Huang, Chen, Wang, Pan, and Wang]{liu2022animating}
Yu Liu, Huai Chen, Lianghua Huang, Di Chen, Bin Wang, Pan Pan, and Lisheng Wang.
\newblock Animating images to transfer clip for video-text retrieval.
\newblock In \emph{ACM SIGIR}, 2022{\natexlab{a}}.

\bibitem[Liu et~al.(2022{\natexlab{b}})Liu, Xiong, Xu, Cao, and Jin]{liu2022ts2}
Yuqi Liu, Pengfei Xiong, Luhui Xu, Shengming Cao, and Qin Jin.
\newblock Ts2-net: Token shift and selection transformer for text-video retrieval.
\newblock In \emph{ECCV}, 2022{\natexlab{b}}.

\bibitem[Loshchilov and Hutter(2016)]{loshchilov2016sgdr}
Ilya Loshchilov and Frank Hutter.
\newblock Sgdr: Stochastic gradient descent with warm restarts.
\newblock \emph{arXiv preprint arXiv:1608.03983}, 2016.

\bibitem[Loshchilov and Hutter(2017)]{loshchilov2017decoupled}
Ilya Loshchilov and Frank Hutter.
\newblock Decoupled weight decay regularization.
\newblock \emph{arXiv preprint arXiv:1711.05101}, 2017.

\bibitem[Luo et~al.(2022)Luo, Ji, Zhong, Chen, Lei, Duan, and Li]{luo2022clip4clip}
Huaishao Luo, Lei Ji, Ming Zhong, Yang Chen, Wen Lei, Nan Duan, and Tianrui Li.
\newblock Clip4clip: An empirical study of clip for end to end video clip retrieval and captioning.
\newblock In \emph{Neurocomputing}, 2022.

\bibitem[Miech et~al.(2018)Miech, Laptev, and Sivic]{miech2018learning}
Antoine Miech, Ivan Laptev, and Josef Sivic.
\newblock Learning a text-video embedding from incomplete and heterogeneous data.
\newblock \emph{arXiv}, 2018.

\bibitem[Oord et~al.(2018)Oord, Li, and Vinyals]{oord2018representation}
Aaron van~den Oord, Yazhe Li, and Oriol Vinyals.
\newblock Representation learning with contrastive predictive coding.
\newblock \emph{arXiv preprint arXiv:1807.03748}, 2018.

\bibitem[Pei et~al.(2023)Pei, Liu, Li, Shao, Xu, Dai, Lu, and Yan]{10203496}
R. Pei, J. Liu, W. Li, B. Shao, S. Xu, P. Dai, J. Lu, and Y. Yan.
\newblock Clipping: Distilling clip-based models with a student base for video-language retrieval.
\newblock In \emph{CVPR}, 2023.

\bibitem[Portillo-Quintero et~al.(2021)Portillo-Quintero, Ortiz-Bayliss, and Terashima-Mar{\'\i}n]{portillo2021straightforward}
Jes{\'u}s~Andr{\'e}s Portillo-Quintero, Jos{\'e}~Carlos Ortiz-Bayliss, and Hugo Terashima-Mar{\'\i}n.
\newblock A straightforward framework for video retrieval using clip.
\newblock In \emph{MCPR}, 2021.

\bibitem[Radford et~al.(2021)Radford, Kim, Hallacy, Ramesh, Goh, Agarwal, Sastry, Askell, Mishkin, Clark, et~al.]{radford2021learning}
Alec Radford, Jong~Wook Kim, Chris Hallacy, Aditya Ramesh, Gabriel Goh, Sandhini Agarwal, Girish Sastry, Amanda Askell, Pamela Mishkin, Jack Clark, et~al.
\newblock Learning transferable visual models from natural language supervision.
\newblock In \emph{ICML}, 2021.

\bibitem[Rohrbach et~al.(2015)Rohrbach, Rohrbach, Tandon, and Schiele]{rohrbach2015dataset}
Anna Rohrbach, Marcus Rohrbach, Niket Tandon, and Bernt Schiele.
\newblock A dataset for movie description.
\newblock In \emph{CVPR}, 2015.

\bibitem[Sigurdsson et~al.(2016)Sigurdsson, Varol, Wang, Farhadi, Laptev, and Gupta]{sigurdsson2016hollywood}
Gunnar~A Sigurdsson, G{\"u}l Varol, Xiaolong Wang, Ali Farhadi, Ivan Laptev, and Abhinav Gupta.
\newblock Hollywood in homes: Crowdsourcing data collection for activity understanding.
\newblock In \emph{ECCV}, 2016.

\bibitem[Vaswani et~al.(2017)Vaswani, Shazeer, Parmar, Uszkoreit, Jones, Gomez, Kaiser, and Polosukhin]{vaswani2017attention}
Ashish Vaswani, Noam Shazeer, Niki Parmar, Jakob Uszkoreit, Llion Jones, Aidan~N Gomez, {\L}ukasz Kaiser, and Illia Polosukhin.
\newblock Attention is all you need.
\newblock \emph{NeurIPS}, 2017.

\bibitem[Wang et~al.(2022{\natexlab{a}})Wang, Zhang, Zheng, Pan, and Hua]{wang2022disentangled}
Qiang Wang, Yanhao Zhang, Yun Zheng, Pan Pan, and Xian-Sheng Hua.
\newblock Disentangled representation learning for text-video retrieval.
\newblock \emph{arXiv}, 2022{\natexlab{a}}.

\bibitem[Wang et~al.(2021{\natexlab{a}})Wang, Zhang, Chen, Cai, Zhou, Peng, Guo, Wu, and Sun]{wang2021dig}
Wenzhe Wang, Mengdan Zhang, Runnan Chen, Guanyu Cai, Penghao Zhou, Pai Peng, Xiaowei Guo, Jian Wu, and Xing Sun.
\newblock Dig into multi-modal cues for video retrieval with hierarchical alignment.
\newblock In \emph{IJCAI}, 2021{\natexlab{a}}.

\bibitem[Wang et~al.(2019)Wang, Wu, Chen, Li, Wang, and Wang]{wang2019vatex}
Xin Wang, Jiawei Wu, Junkun Chen, Lei Li, Yuan-Fang Wang, and William~Yang Wang.
\newblock Vatex: A large-scale, high-quality multilingual dataset for video-and-language research.
\newblock In \emph{ICCV}, 2019.

\bibitem[Wang et~al.(2021{\natexlab{b}})Wang, Zhu, and Yang]{wang2021t2vlad}
Xiaohan Wang, Linchao Zhu, and Yi Yang.
\newblock T2vlad: global-local sequence alignment for text-video retrieval.
\newblock In \emph{CVPR}, 2021{\natexlab{b}}.

\bibitem[Wang et~al.(2022{\natexlab{b}})Wang, Chen, Hu, and Li]{LUNVR}
Ziyue Wang, Aozhu Chen, Fan Hu, and Xirong Li.
\newblock Learn to understand negation in video retrieval.
\newblock In \emph{ACM MM}, 2022{\natexlab{b}}.

\bibitem[Wu et~al.(2023)Wu, Luo, Fang, Wang, and Ouyang]{wu2023cap4video}
Wenhao Wu, Haipeng Luo, Bo Fang, Jingdong Wang, and Wanli Ouyang.
\newblock Cap4video: What can auxiliary captions do for text-video retrieval?
\newblock In \emph{CVPR}, 2023.

\bibitem[Xu et~al.(2021)Xu, Ghosh, Huang, Okhonko, Aghajanyan, Metze, Zettlemoyer, and Feichtenhofer]{xu-etal-2021-videoclip}
Hu Xu, Gargi Ghosh, Po-Yao Huang, Dmytro Okhonko, Armen Aghajanyan, Florian Metze, Luke Zettlemoyer, and Christoph Feichtenhofer.
\newblock {V}ideo{CLIP}: Contrastive pre-training for zero-shot video-text understanding.
\newblock In \emph{EMNLP}, 2021.

\bibitem[Xu et~al.(2016)Xu, Mei, Yao, and Rui]{xu2016msr}
Jun Xu, Tao Mei, Ting Yao, and Yong Rui.
\newblock Msr-vtt: A large video description dataset for bridging video and language.
\newblock In \emph{CVPR}, 2016.

\bibitem[Xue et~al.(2022)Xue, Hang, Zeng, Sun, Liu, Yang, Fu, and Guo]{xue2022advancing}
Hongwei Xue, Tiankai Hang, Yanhong Zeng, Yuchong Sun, Bei Liu, Huan Yang, Jianlong Fu, and Baining Guo.
\newblock Advancing high-resolution video-language representation with large-scale video transcriptions.
\newblock In \emph{CVPR}, 2022.

\bibitem[Xue et~al.(2023)Xue, Sun, Liu, Fu, Song, Li, and Luo]{xue2022clip}
Hongwei Xue, Yuchong Sun, Bei Liu, Jianlong Fu, Ruihua Song, Houqiang Li, and Jiebo Luo.
\newblock Clip-vip: Adapting pre-trained image-text model to video-language representation alignment.
\newblock In \emph{ICLR}, 2023.

\bibitem[Yu et~al.(2018)Yu, Kim, and Kim]{yu2018joint}
Youngjae Yu, Jongseok Kim, and Gunhee Kim.
\newblock A joint sequence fusion model for video question answering and retrieval.
\newblock In \emph{ECCV}, 2018.

\bibitem[Zhao et~al.(2023)Zhao, Jiao, Xie, and Lin]{10209009}
N. Zhao, J. Jiao, W. Xie, and D. Lin.
\newblock Cali-nce: Boosting cross-modal video representation learning with calibrated alignment.
\newblock In \emph{CVPRW}, 2023.

\bibitem[Zhao et~al.(2022)Zhao, Zhu, Wang, and Yang]{zhao2022centerclip}
Shuai Zhao, Linchao Zhu, Xiaohan Wang, and Yi Yang.
\newblock Centerclip: Token clustering for efficient text-video retrieval.
\newblock In \emph{ACM SIGIR}, 2022.

\bibitem[Zhu and Yang(2020)]{zhu2020actbert}
Linchao Zhu and Yi Yang.
\newblock Actbert: Learning global-local video-text representations.
\newblock In \emph{CVPR}, 2020.

\end{thebibliography}
}


\end{document}